\documentclass[journal]{IEEEtran}

\IEEEoverridecommandlockouts

\usepackage{multirow}
\usepackage{graphicx}
\usepackage{comment}
\usepackage{cite}
\usepackage{amsmath}
\usepackage{amsfonts}
\usepackage{mathtools}
\usepackage{tabularx}
\usepackage{graphicx}
\usepackage{array}
\usepackage{float}
\usepackage{hhline}
\usepackage{algorithmic}
\usepackage{esvect}
\usepackage[linesnumbered, ruled, vlined]{algorithm2e}
\usepackage[colorlinks,urlcolor=blue]{hyperref}
\usepackage{setspace}
\usepackage{colortbl}
\usepackage{arydshln}
\usepackage{nicefrac}
\usepackage{float}
\usepackage{hhline}
\SetCommentSty{mycommfont}

\DeclareMathOperator*{\argmin}{arg\,min}

\title{Direct LiDAR-Inertial Odometry and Mapping: \\
Perceptive and Connective SLAM}

\author{Kenny Chen$^{1}$, Ryan Nemiroff$^{1}$, and Brett T. Lopez$^{2}$%
\thanks{$^{1}$Kenny Chen and Ryan Nemiroff are with the Department of Electrical and Computer Engineering, University of California Los Angeles, Los Angeles, CA, USA. {\tt\footnotesize \{kennyjchen, ryguyn\}@ucla.edu}}%
\thanks{$^{2}$Brett T. Lopez is with the Department of Mechanical and Aerospace Engineering, University of California Los Angeles, Los Angeles, CA, USA. {\tt\footnotesize btlopez@ucla.edu}}%
\thanks{All authors are with the Verifiable and Control-Theoretic Robotics Laboratory, University of California Los Angeles, Los Angeles,
CA, USA.}%
}

\begin{document}
\maketitle


\begin{abstract}
This paper presents Direct LiDAR-Inertial Odometry and Mapping (DLIOM), a robust SLAM algorithm with an explicit focus on computational efficiency, operational reliability, and real-world efficacy.
DLIOM contains several key algorithmic innovations in both the front-end and back-end subsystems to design a resilient LiDAR-inertial architecture that is perceptive to the environment and produces accurate localization and high-fidelity 3D mapping for autonomous robotic platforms.
Our ideas spawned after a deep investigation into modern LiDAR SLAM systems and their inabilities to generalize across different operating environments, in which we address several common algorithmic failure points by means of proactive safe-guards to provide long-term operational reliability in the unstructured real world.
We detail several important innovations to localization accuracy and mapping resiliency distributed throughout a typical LiDAR SLAM pipeline to comprehensively increase algorithmic speed, accuracy, and robustness.
In addition, we discuss insights gained from our ground-up approach while implementing such a complex system for real-time state estimation on resource-constrained systems, and we experimentally show the increased performance of our method as compared to the current state-of-the-art on both public benchmark and self-collected datasets.
\end{abstract}


\section{Introduction}
\label{sec:introduction}

Accurate real-time state estimation and mapping are fundamental capabilities that are the backbone for autonomous robots to perceive, plan, and navigate through unknown environments. 
Long-term operational reliability of such capabilities require algorithmic resiliency against off-nominal conditions, such as the presence of particulates (e.g., dust or fog), low-lighting, difficult or unstructured landscape, and other external factors.
While visual SLAM approaches may work in well-lit environments, they quickly break down in-the-wild from their strong environmental assumptions, brittle architecture, or high computational complexity. LiDAR-based methods, on the other hand, have recently become a viable option for many mobile platforms due to lighter and cheaper sensors.
As a result, researchers have recently developed several new LiDAR odometry (LO) and LiDAR-inertial odometry (LIO) systems which often outperform vision-based localization due to the sensor's superior range and depth measurement accuracy.
However, there are still several fundamental challenges in developing robust, long-term LiDAR-centric SLAM solutions, especially for autonomous robots that explore unknown environments, execute agile maneuvers, or traverse uneven terrain \cite{cadena2016past}.

\begin{figure}[!t]
    \centering
    \includegraphics[width=0.95\columnwidth]{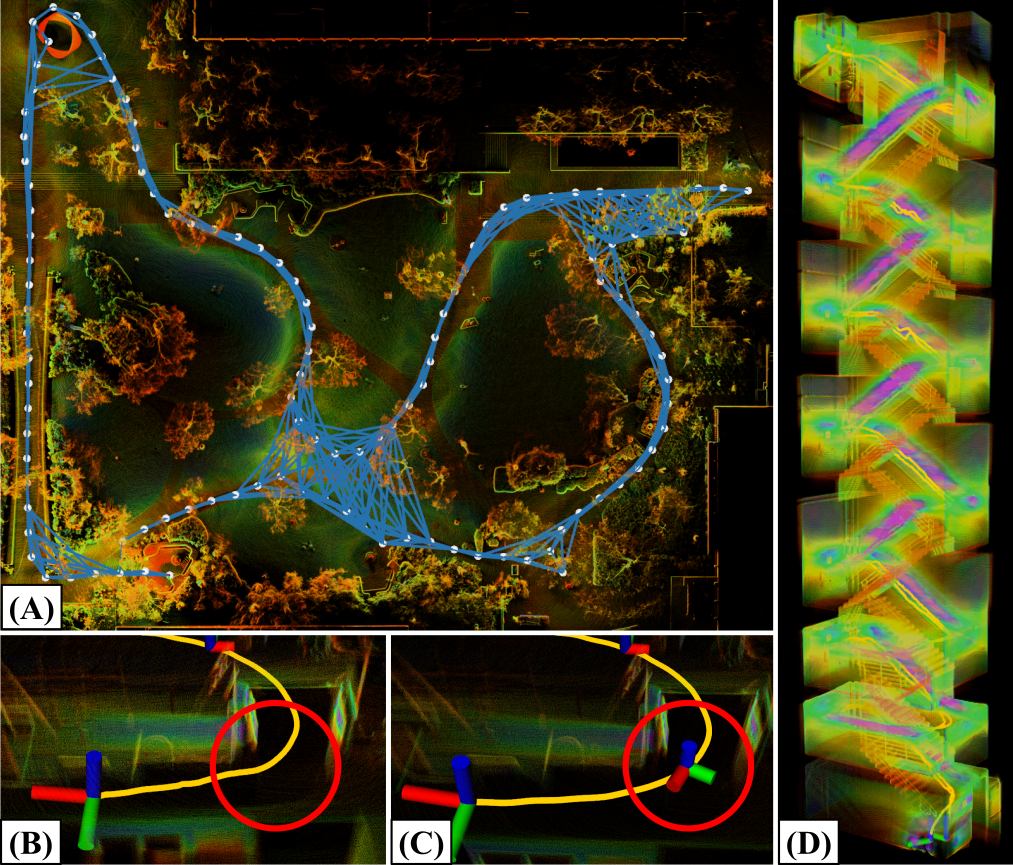}
    \caption{\textbf{Dense Connective Mapping with Resilient Localization.} Our novel DLIOM algorithm contains several proactive safe-guards against common failure points in LiDAR odometry to create a resilient SLAM framework that adapts to its operating environment. (A) A top-down view of UCLA's Sculpture Garden mapped by DLIOM, showcasing the algorithm's derived pose graph with interkeyframe constraints for local accuracy and global resiliency. (B \& C) An example of DLIOM's slip-resistant keyframing which helps anchor scan-to-map registration, in which abrupt scenery changes (e.g., traversal through a door) that normally cause slippage (B) are mitigated by scene change detection (C). (D) A map of an eight-story staircase generated by DLIOM, showcasing the difficult environments our algorithm can track in.}
    \label{fig:main}
    \vskip -0.2in
\end{figure}

Algorithmic resiliency by means of building proactive safe-guards against common failure points in SLAM can provide perceptive and failure-tolerant localization across a wide range of operating environments for long-term reliability. While existing algorithms may work well in structured environments that pose well-constrained problems for the back-end optimizer, their performance can quickly degrade under irregular conditions---yielding slow, brittle perception systems unsuitable for real-world use. Of the few recent approaches that do have the ability to adapt to different conditions on-the-fly, they either rely on switching to other sensing modalities in degraded environments \cite{tagliabue2020lion}, require complex parameter tuning procedures based on manually specified heuristics \cite{koide2021adaptive, koide2021automatic}, or focus solely on the scan-matching process in an effort to better anchor weakly-constrained registration problems \cite{tuna2022x}. 

To this end, this paper presents the Direct LiDAR-Inertial Odometry and Mapping (DLIOM) algorithm, a robust, real-time SLAM system with several key innovations that provide increased resiliency and accuracy for both localization and mapping (Fig.~\ref{fig:main}). The main contributions of this work are five-fold, each targetting a specific module in a typical LiDAR SLAM architecture (bolded) to comprehensively increase algorithmic speed, accuracy, and robustness:
\begin{itemize}
    \item \textbf{Keyframing}: A method for slip-resistant keyframing by detecting the onset of scan-matching slippage during abrupt scene changes via a global and sensor-agnostic degeneracy metric.
    \item \textbf{Submapping}: A method which generates explicitly-relevant local submaps with maximum coverage by computing the relative 3D Jaccard index for each keyframe for scan-to-map registration.
    \item \textbf{Mapping}: A method to increase local mapping accuracy and global loop closure resiliency via connectivity factors and keyframe-based loop closures.
    \item \textbf{Scan-Matching}: An adaptive scan-matching method via a novel point cloud sparsity metric for consistent registration in both large and small environments.
    \item \textbf{Motion Correction}: A new coarse-to-fine technique for fast and parallelizable point-wise motion correction, in which a set of analytical equations with a constant jerk and angular acceleration motion model is derived for constructing continuous-time trajectories.
\end{itemize}

This paper substantially extends algorithmically and experimentally our previous work, Direct LiDAR-Inertial Odometry (DLIO)~\cite{chen2022dlio}. 
In particular, aside from the continuous-time motion correction, each listed contribution above is an original key idea in order to increase algorithmic accuracy, generalization, and failure resiliency. 
Moreover, we provide new algorithmic insights from our ground-up approach, in addition to new experimental results for algorithmic resiliency, computational efficiency, and overall map and trajectory accuracy as compared to the state-of-the-art. 

\begin{figure*}[!t]
    \centering
    \includegraphics[width=0.99\textwidth]{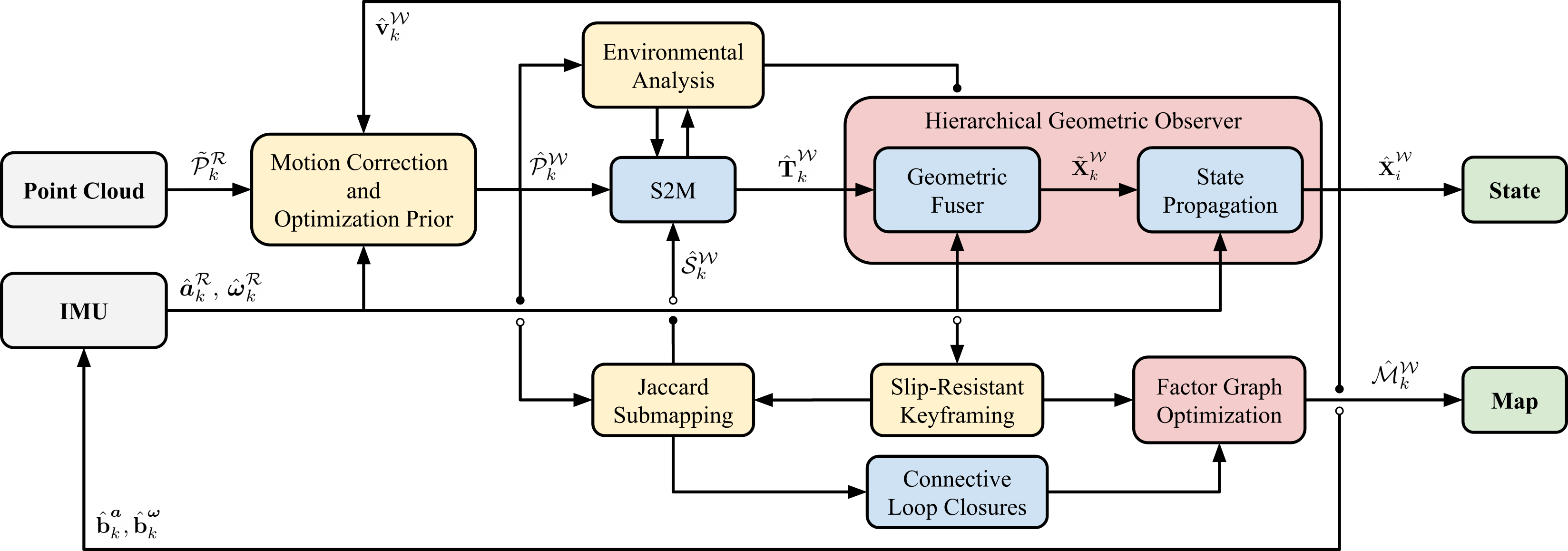}
    \caption{\textbf{System Architecture.} DLIOM's two-pronged architecture contains several key innovations to provide a comprehensive SLAM pipeline with real-world operational reliability. Point-wise continuous-time integration in $\mathcal{W}$ ensures maximum fidelity of the corrected cloud and is registered onto the robot's map by a custom GICP-based scan-matcher. An analysis on the environmental structure and health of scan-matching provides several system metrics for adaptively tuning maximum correspondence distance, in addition to slip-resistant keyframing. Additionally, a 3D Jaccard index for each keyframe is computed against the current scan to maximize submap coverage and therefore scan-matching correspondences. The system's state is subsequently updated by a nonlinear geometric observer with strong convergence properties, and these estimates of pose, velocity, and bias then initialize the next iteration. This system state is also subsequently sent to a background mapping thread, which places pose graph nodes at keyframe locations and builds a connective graph via interkeyframe constraints for local accuracy and global resiliency.}
    \label{fig:architecture}
\end{figure*}

\section{Related Work}
\label{sec:relatedwork}

Simultaneous localization and mapping (SLAM) algorithms for 3D time-of-flight sensors (e.g., LiDAR) rely on aligning point clouds by solving a nonlinear least-squares problem that minimizes the error across corresponding points and/or planes. To find point/plane correspondences, popular methods such as the iterative closest point (ICP) algorithm \cite{besl1992method, chen1992object}, Generalized-ICP (GICP) \cite{segal2009generalized}, or the normal distribution transform~\cite{ndt} recursively match potential corresponding entities until alignment converges to a local minimum. 
Slow convergence time is often observed when determining correspondences for a large set of points, so 
\textit{feature}-based methods \cite{zhang2014loam, shan2018lego, shan2020lio, shan2021lvi, pan2021mulls, xu2021fast, nguyen2021miliom, ye2019tightly} attempt to extract only the most salient data points, e.g., corners and edges, in a scan to decrease computation time. However, the efficacy of feature extraction is highly dependent on specific implementation. Moreover, useful points are often discarded resulting in inaccurate estimates and maps. Conversely, \textit{dense} methods \cite{palieri2020locus, tagliabue2020lion, chen2022direct, xu2022fast, reinke2022iros} directly align acquired scans but often rely heavily on aggressive voxelization---a process that can alter important data correspondences---to achieve real-time performance.

LiDAR odometry approaches can also be broadly classified according to their method of incorporating other sensing modalities into the estimation pipeline.
\textit{Loosely}-coupled methods \cite{zhang2014loam, shan2018lego, palieri2020locus, tagliabue2020lion, chen2022direct} process data sequentially.
For example, IMU measurements are used to augment LiDAR scan registration by providing an optimization prior. 
These methods are often quite robust due to the precision of LiDAR measurements, but localization results can be less accurate as only a subset of all available data is used for estimation.
\textit{Tightly}-coupled methods \cite{shan2020lio, xu2022fast, ye2019tightly, nguyen2021miliom, wang2022dliom}, on the otherhand, can offer improved accuracy by jointly considering measurements from all sensing modalities.
These methods commonly employ either graph-based optimization \cite{shan2020lio, ye2019tightly, nguyen2021miliom, zhang2016degeneracy} or a stochastic filtering framework, e.g., Kalman filter \cite{xu2021fast, xu2022fast}. However, compared to geometric observers \cite{baldwin2007complementary, vasconcelos2008nonlinear}, these approaches possess minimal convergence guarantees even in the most ideal settings which can result in significant localization error from inconsistent sensor fusion and map deformation from incorrect scan placement.

Incorporating additional sensors can also aid in correcting motion-induced point cloud distortion. 
For example, LOAM~\cite{zhang2014loam} compensates for spin distortion by iteratively estimating sensor pose via scan-matching and a loosely-coupled IMU using a constant velocity assumption. Similarly, LIO-SAM~\cite{shan2020lio} formulates LiDAR-inertial odometry atop a factor graph to jointly optimize for body velocity, and in their implementation, points were subsequently deskewed by linearly interpolating rotational motion. FAST-LIO~\cite{xu2021fast} and FAST-LIO2~\cite{xu2022fast} instead employ a back-propagation step on point timestamps after a forward-propagation of IMU measurements to produce relative transformations to the scan-end time. However, these methods (and others \cite{renzler2020increased, deschenes2021lidar}) all operate in \textit{discrete}-time which may induce a loss in precision, leading to a high interest in \textit{continuous}-time methods. Elastic LiDAR Fusion~\cite{park2018elastic}, for example, handles scan deformation by optimizing for a continuous linear trajectory, whereas Wildcat~\cite{ramezani2022wildcat} and \cite{droeschel2018efficient} instead iteratively fit a cubic B-spline to remove distortion from surfel maps. More recently, CT-ICP~\cite{dellenbach2022ct} and ElasticLiDAR++~\cite{park2022elasticity} use a LiDAR-only approach to define a continuous-time trajectory parameterized by two poses per scan, which allows for elastic registration of the scan during optimization. However, these methods can still be too simplistic in modeling the trajectory under highly dynamical movements or may be too computationally costly to work reliably in real-time.

Another crucial component in LIO systems is submapping, which involves extracting a smaller point cloud from the global map for efficient processing and to increase pose estimation consistency. Rather than processing the entire map on every iteration which is often computationally intractable, a \textit{submap} instead contains only a subset of all available data points to be considered. However, the efficacy of a submapping strategy depends on its ability to extract only the most relevant map points for scan-matching to avoid any wasted computation when constructing corresponding data structures (i.e., kdtrees, normals). One common approach is to use a sliding window such that the submap consists of a set of recent scans \cite{ye2019tightly, liu2021balm}. However, this method assumes a strong temporal correspondence between points which may not always be the case (i.e., revisiting a location) and may not perform well under significant drift. To mitigate this, radius-based approaches \cite{palieri2020locus, shan2020lio} extract points nearby the current position by directly working with point clouds and continually adding points to an internal octree data structure. This, however, results in unbounded growth in the map \cite{vizzo2023kiss} and therefore an explosion in computational expense through the large number of nearest neighbor calculations required, which is infeasible for real-time usage. Keyframe-based methods \cite{shan2020lio, chen2022direct}, on the other hand, link keyed locations in space to its corresponding scan, and therefore reduces the search space required to extract a comprehensive submap. However, previous methods \cite{chen2022direct, chen2022dlio} have still implicitly assumed that nearby keyframes contain the most relevant data points for a submap and do not explicitly compute a metric of relevancy per keyframe, risking the extraction of keyframes which may not be used at all.

More recently, researchers have been interested in building new methods of algorithmic resiliency into odometry pipelines to ensure reliable localization across a diverse set of environments. While the field of adaptive localization is still in its infancy, early works have pioneered the idea of adaptivity in unstructured and/or extreme environments. For instance, to provide resiliency against LiDAR slippage in geometrically-degenerate environments, LION \cite{tagliabue2020lion}, DARE-SLAM \cite{ebadi2021dare}, and DAMS-LIO \cite{jiao2023dams} proposed using the condition number of the scan-matching Hessian as an observability score in order to switch to a different state-estimation algorithm (e.g., visual-inertial). However, these works assume the availability of other odometry paradigms on-board which may not always be available. On the other hand, works such as \cite{koide2021adaptive} and \cite{koide2021automatic} attempt to automatically tune system hyperparameters based on trajectory error modeling, but these require an expensive offline training procedure to retrieve optimal parameter values and do not generalize to other environments outside of the training set. Similarly, KISS-ICP \cite{vizzo2023kiss} adaptively tunes maximum correspondence distance for ICP, but their method scales the metric according to robot acceleration which does not generalize across differently-sized environments. AdaLIO~\cite{lim2023adalio}, on the other hand, automatically tunes voxelization, search distance, and plane residual parameters in detected degenerate cases. Methods of global loop closure detection can also aid in reducing drift in the map through place recognition and relocalization, using descriptors and detectors such as ScanContext~\cite{gkim-2018-iros, gskim-2021-tro} and Segregator~\cite{yin2023segregator}. More recently, \cite{zhang2016degeneracy} and X-ICP\cite{tuna2022x} propose innovative online methods to mitigate degeneracy by analyzing the geometric structure of the optimization constraints in scan-to-scan and scan-to-map, respectively, but these works only target a specific module in an entire, complex SLAM pipeline.

To this end, DLIOM proposes several new techniques to LiDAR-based SLAM systems which address several deficiencies in both the front-end and back-end. Our ideas target different scales in the data processing pipeline to progressively increase localization resiliency and mapping accuracy. First, a fast, coarse-to-fine approach constructs continuous-time trajectories between each LiDAR sweep for accurate, parallelizable point-wise motion correction. These motion-corrected clouds are then incrementally registered against an extracted submap via an adaptive scan-matching technique, which tunes the maximum correspondence distance based on the current cloud's sparsity for consistent registration across different environments. Each extracted submap is explicitly generated by computing each environmental keyframe's relevancy towards the current scan-matching problem via a relative 3D Jaccard index; this is done to maximize submap coverage and therefore data association between the scan and submap. To prevent slippage, scan-matching health is continually monitored through a novel sensor-agnostic degeneracy metric, which inserts a new keyframe when optimization is too weakly-constrained during rapid scene changes. Finally, to increase local mapping accuracy and global loop closure resiliency, we compute interkeyframe overlap to provide additional factors to our keyframe-based factor graph mapper.

\section{System Overview \& Data Processing}
\label{sec:methods}

DLIOM is a robust SLAM algorithm with a specific focus on localization resiliency, mapping accuracy, and real-world operational reliability~(Fig.~\ref{fig:architecture}). The architecture contains two parallel threads which process odometry estimation and global mapping in real-time. In the first, LiDAR scans are consecutively motion-corrected and then registered against a keyframe-based submap to provide an accurate update signal for integrated IMU measurements. This accuracy is ensured by an adaptive maximum correspondence distance for consistent scan-matching, in addition to how we explicitly derive the local submap, whereby maximum submap coverage of the current scan is enforced by computing a 3D Jaccard index for each environmental keyframe. This helps increase robustness against errors in data association during GICP optimization. A novel method for computing environmental degeneracy provides a global notion of scan-matching slippage and continually monitors optimization health status, placing a new keyframe right before the onset of slippage, and a nonlinear geometric observer fuses LiDAR scan-matching and IMU preintegration to provide high-rate state estimation with certifiable convergence guarantees.

In the second thread, keyframes continually build upon an internal factor graph, whereby each keyframe is represented as a node in the graph and various constraints between pairs of keyframes are factors between nodes. ``Sequential" factors between adjacent keyframes provide a strong backbone to the pose graph and is feasible due to our system's accurate local odometry; ``connective" factors between overlapping keyframes (via their 3D Jaccard index) provide local map accuracy and global map resiliency against catastrophically incorrect loop closures. Frame offsets after such loop closures are carefully managed in order to prevent discontinuities in estimated velocity for safe robot control.
Our algorithm is completely built from the ground-up to decrease computational overhead and increase algorithmic failure-tolerance and real-world reliability.

\begin{algorithm}[!tb]
    \setstretch{1}
	\SetAlgoLined
	\textbf{input:} $\hat{\textbf{X}}_{k\text{-}1}^{\mathcal{W}}$, $\hat{\mathcal{M}}_k^{\mathcal{W}}$, $\mathcal{P}_k^{\mathcal{L}}$, $\boldsymbol{a}_{k}^{\mathcal{B}}$,  $\boldsymbol{\omega}_{k}^{\mathcal{B}}$; \,
	\textbf{output:} $\hat{\textbf{X}}_i^{\mathcal{W}}$ \\
	
	\BlankLine
	\small {\tcp{LiDAR Callback Thread}}
	\While{$\mathcal{P}_k^{\mathcal{L}} \neq \emptyset$} {

        \BlankLine
	    \small {\tcp{initialize points and transform to $\mathcal{R}$}}
	    $\mathcal{\tilde{P}}_k^\mathcal{R}$ $\leftarrow$ initializePointCloud$(\,\mathcal{P}_k^\mathcal{L}\,)$; (\ref{sec:preprocessing})\\

        \BlankLine
        \small {\tcp{construct coarse trajectory}}
        \For{$\hat{\boldsymbol{a}}_i^{\mathcal{R}}, \hat{\boldsymbol{\omega}}_i^{\mathcal{R}}$ between $t_{k\text{-}1}$ and $t_{k}$} {
            $\hat{\textbf{p}}_{i}, \hat{\textbf{v}}_{i}, \hat{\textbf{q}}_{i} \leftarrow$ discreteInt$(\,\hat{\textbf{X}}_{k\text{-}1}^{\mathcal{W}},\,\hat{\boldsymbol{a}}_{i\text{-}1}^{\mathcal{R}} ,\, \hat{\boldsymbol{\omega}}_{i\text{-}1}^{\mathcal{R}}\,); (\ref{eq:deskew})$ \\
            $\hat{\textbf{T}}_{i}^{\mathcal{W}} = [ \, \hat{\textbf{R}}(\hat{\textbf{q}}_{i}) \, | \, \hat{\textbf{p}}_{i} \, ]$;\\
        }

        \BlankLine
        \small {\tcp{continuous-time motion correction}}
        \For{$p_k^n \in \mathcal{\tilde{P}}_k^\mathcal{R}$} {
            $\hat{\textbf{T}}^{\mathcal{W}*}_{n} \leftarrow$ continuousInt$(\,\hat{\textbf{T}}^{\mathcal{W}*}_{i},\, t_n\,); (\ref{eq:deskew_timestamp})$ \\
            $\hat{p}_k^n = \hat{\textbf{T}}^{\mathcal{W}*}_{n} \otimes p_k^n \,$; 
            $\hat{\mathcal{P}}_k^\mathcal{W}$.append$(\,\hat{p}_k^n\,)$;\\
        }

        \BlankLine
        \small {\tcp{environmental analysis: \\ compute spaciousness and cloud sparsity}}
        $m_k, z_k \leftarrow$ computeAdaptiveParams$(\,\hat{\mathcal{P}}_k^\mathcal{W}\,)$; \cite{chen2022direct}, (\ref{eq:sparsity})

        \BlankLine
	    \small {\tcp{construct submap via 3D Jaccard index}}
        \For{$\mathcal{K}_j^{\mathcal{W}} \in \hat{\mathcal{M}}_k^{\mathcal{W}}$} {
            $J(\hat{\mathcal{P}}_k^{\mathcal{W}}, \mathcal{K}_j^{\mathcal{W}}) \leftarrow {|\mathcal{\hat{\mathcal{P}}}_k^{\mathcal{W}} \cap \mathcal{K}_j^{\mathcal{W}}|} \, / \, {|\mathcal{\hat{\mathcal{P}}}_k^{\mathcal{W}} \cup \mathcal{K}_j^{\mathcal{W}}|}$; (\ref{eq:jaccard})\\
            \If {$J(\hat{\mathcal{P}}_k^{\mathcal{W}}, \mathcal{K}_j^{\mathcal{W}}) \geq thresh_{\text{jaccard}}$} {
                $\hat{\mathcal{S}}_k^\mathcal{W}$.append$(\, \hat{\mathcal{P}}_k^{\mathcal{W}} \,)$; \\
            }
        }

        \BlankLine
        \small {\tcp{adaptive scan-to-map registration}}
	    $\hat{\textbf{T}}_k^{\mathcal{W}} \leftarrow$ GICP$(\,\hat{\mathcal{P}}_k^\mathcal{W},\, \hat{\mathcal{S}}_k^\mathcal{W},\, z_k) \,$; (\ref{eq:gicp})\\

        \BlankLine
        \small {\tcp{slip-resistant keyframing}}
	    \If{$\hat{\mathcal{P}}_k^\mathcal{W}$ is a keyframe via (\ref{eq:degeneracy}) or \cite{chen2022direct}} {
     
            \BlankLine
            $\mathcal{K}_k^{\mathcal{W}} \leftarrow \hat{\mathcal{P}}_k^\mathcal{W}$; \\

            \BlankLine
            \small {\tcp{compute new keyframe overlap and \\ append to connectivity matrix}}
            \For{$\mathcal{K}_j^{\mathcal{W}} \in \hat{\mathcal{M}}_k^{\mathcal{W}}$} {
                $\mathbf{C}_{kj} \leftarrow J(\mathcal{K}_k^{\mathcal{W}}, \mathcal{K}_{j}^{\mathcal{W}})$; (\ref{sec:connectivity})\\
            }
            
            \BlankLine
            \small {\tcp{send new keyframe and updated \\ connectivity matrix to mapping thread}}
            odom2map$(\, \mathcal{K}_k^\mathcal{W}, \mathbf{C}\,)$; \\
     
	    }

        \BlankLine
	    \small {\tcp{geometric observer: state update}}
	    $\hat{\textbf{X}}_k^\mathcal{W} \leftarrow$ updateState$(\,\hat{\textbf{T}}_k^{\mathcal{W}},\, \Delta t^{+}_k\,) \,$; (\ref{sec:methods:geo})\\

        \BlankLine    
	    \Return $\hat{\textbf{X}}_k^{\mathcal{W}}$ \\
	    
	}
	
	\BlankLine
	\small {\tcp{IMU Callback Thread}}
	\While{$\boldsymbol{a}_i^{\mathcal{B}} \neq \emptyset$ \textbf{and} $\boldsymbol{\omega}_i^{\mathcal{B}} \neq \emptyset$} {

        \BlankLine
	    \small {\tcp{apply biases and transform to $\mathcal{R}$}}
	    $\hat{\boldsymbol{a}}_i^{\mathcal{R}} ,\, \hat{\boldsymbol{\omega}}_i^{\mathcal{R}}$ $\leftarrow$ initializeImu$(\,\boldsymbol{a}_i^{\mathcal{B}} ,\, \boldsymbol{\omega}_i^{\mathcal{B}}\,)$; (\ref{sec:preprocessing})\\

        \BlankLine
	    \small {\tcp{geometric observer: state propagation}}
	    $\hat{\textbf{X}}_i^{\mathcal{W}} \leftarrow$ propagateState$(\hat{\textbf{X}}_{k}^{\mathcal{W}}, \hat{\boldsymbol{a}}_{i}^{\mathcal{R}}, \hat{\boldsymbol{\omega}}_{i}^{\mathcal{R}}, \Delta t^{+}_i)$; (\ref{sec:methods:geo})\\

        \BlankLine
	    \Return $\hat{\textbf{X}}_i^{\mathcal{W}}$ \\
	    
	}

	\caption{DLIOM: Odometry Thread}
	\label{alg:state_estimation}
	
\end{algorithm}

\begin{algorithm}[!tb]
    \setstretch{1}
	\SetAlgoLined
	\textbf{input:} $\mathcal{K}_k^\mathcal{W}$, $\mathbf{C}$, $\mathcal{G}_k$; \,
	\textbf{output:} $\hat{\mathcal{M}}_k^{\mathcal{W}}$ \\
	\textbf{initialize:} $\mathcal{G}_k$.addPriorFactor$(\,\mathcal{K}_0^\mathcal{W}\,)$

        \BlankLine
        \small {\tcp{Keyframe Callback Thread}}
	\While{$\mathcal{K}_k^\mathcal{W} \neq \emptyset$ \textbf{and} $\mathbf{C} \neq \emptyset$} {
        \BlankLine
    	\small {\tcp{add factor to previous keyframe}}
            $\mathcal{G}_k$.addBetweenFactor$(\, \mathcal{K}_{k-1}^\mathcal{W}, \mathcal{K}_k^\mathcal{W} \,)$; \\
     
        \BlankLine
    	\small {\tcp{add factors to connective keyframes}}
            \For{$\mathcal{K}_j^{\mathcal{W}} \in \hat{\mathcal{M}}_k^{\mathcal{W}}$}{
                \If {$\mathbf{C}_{kj} \geq thresh_{\text{conn}}$} {
                    $\mathcal{G}_k$.addBetweenFactor$(\, \mathcal{K}_k^\mathcal{W}, \mathcal{K}_j^\mathcal{W} \,)$;
                }
            }
        \BlankLine
    	\small {\tcp{submap-based loop closures}}
            \For{$\mathcal{K}_j^{\mathcal{W}} \in \hat{\mathcal{M}}_k^{\mathcal{W}}$}{

                \If{$\mathcal{K}_j^{\mathcal{W}}$ is a loop candidate via \ref{sec:loopclosures}} {
                    
                    $\mathcal{L}_k^{\mathcal{W}} \leftarrow \mathcal{L}_k^{\mathcal{W}} \oplus \mathcal{K}_j^{\mathcal{W}}$
                
                }

            }
            
            $\hat{\textbf{T}} \leftarrow$ GICP$(\,\mathcal{L}_k^{\mathcal{R}},\, \hat{\mathcal{K}}_k^\mathcal{R},\, z_k\,)$; (\ref{eq:gicp})\\
            
            \If {$\text{fitnessScore} \,\geq\, thresh_{\text{loop}}$} {
                $\mathcal{G}_k$.addBetweenFactor$(\, \mathcal{K}_k^\mathcal{W}, \mathcal{K}_{\mathcal{L}}^\mathcal{W} \,)$;
            }
    
        \BlankLine
    	\small {\tcp{optimize the factor graph}}
            $\mathcal{G}_k$.optimize$(\, \mathcal{K}_k^{\mathcal{W}} \,)$;

        \BlankLine
            \small {\tcp{update map after optimization}}
            $\hat{\mathcal{M}}_k^{\mathcal{W}} \leftarrow$ updateKeyframes( $\mathcal{G}_k$ );
            
        \BlankLine
    	\small {\tcp{send to odometry thread}}
        map2odom$(\, \hat{\mathcal{M}}_k^{\mathcal{W}} \,)$;
    
        \BlankLine
        \Return $\hat{\mathcal{M}}_k^{\mathcal{W}}$ \\

    }

	\caption{DLIOM: Mapping Thread}
	\label{alg:mapping}
	
\end{algorithm}

\subsection{Mathematical Notation}
Let the point cloud for a single LiDAR sweep initiated at time $t_k$ be denoted as $\mathcal{P}_k$ and indexed by $k$.
The point cloud $\mathcal{P}_k$ is composed of points $p^n_{k} \in \mathbb{R}^3$ that are measured at a time $\Delta t^n_{k}$ relative to the start of the scan and indexed by ${n=1,\hdots,N}$ where $N$ is the total number of points in the scan.
The world frame is denoted as $\mathcal{W}$ and the robot frame as $\mathcal{R}$ located at its center of gravity, with the convention that $x$ points forward, $y$ left, and $z$ up. The IMU's coordinate system is denoted as $\mathcal{B}$ and the LiDAR's as $\mathcal{L}$, and the robot's state vector $\textbf{X}_k$ at index $k$ is defined as the tuple
\begin{equation}
    \label{eq:state}
    \textbf{X}_k = \left[ \, \textbf{p}^{\mathcal{W}}_{k} ,\, \textbf{q}^{\mathcal{W}}_{k} ,\, \textbf{v}^{\mathcal{W}}_{k} ,\, \textbf{b}^{\boldsymbol{a}}_{k} ,\, \textbf{b}^{\boldsymbol{\omega}}_{k} \, \right] ^\top \,,
\end{equation}
\noindent where $\textbf{p}^{\mathcal{W}} \in \mathbb{R}^3$ is the robot's position, $\textbf{q}^{\mathcal{W}}$ is the orientation encoded by a four vector quaternion on $\mathbb{S}^3$ under Hamilton notation, $\textbf{v}^{\mathcal{W}} \in \mathbb{R}^3$ is the robot's velocity, $\textbf{b}^{\boldsymbol{a}} \in \mathbb{R}^3$ is the accelerometer's bias, and $\textbf{b}^{\boldsymbol{\omega}} \in \mathbb{R}^3$ is the gyroscope's bias. Measurements $\hat{\boldsymbol{a}}$ and $\hat{\boldsymbol{\omega}}$ from an IMU are modeled as
\begin{align}
    \label{eq:imu}
    \hat{\boldsymbol{a}}_i &= (\boldsymbol{a}_i - \boldsymbol{g}) + \textbf{b}_i^{\boldsymbol{a}} + \textbf{n}_i^{\boldsymbol{a}} \,, \\
    \hat{\boldsymbol{\omega}}_i &= \boldsymbol{\omega}_i + \textbf{b}_i^{\boldsymbol{\omega}} + \textbf{n}_i^{\boldsymbol{\omega}} \,,
\end{align}
\noindent and indexed by ${i=1,\hdots,M}$ for $M$ measurements between clock times $t_{k\text{-}1}$ and $t_{k}$. With some abuse of notation, indices $k$ and $i$ occur at LiDAR and IMU rate, respectively, and will be written this way for simplicity unless otherwise stated. Raw sensor measurements $\boldsymbol{a}_i$ and $\boldsymbol{\omega}_i$ contain bias $\textbf{b}_i$ and white noise $\textbf{n}_i$, and $\boldsymbol{g}$ is the rotated gravity vector. In this work, we address the following problem: given a distorted point cloud $\mathcal{P}_k$ from a LiDAR and $\boldsymbol{a}_{i}$ and $\boldsymbol{\omega}_{i}$ from an IMU, estimate the robot's state $\hat{\textbf{X}}_i^{\mathcal{W}}$ and the geometric map $\hat{\mathcal{M}}_k^{\mathcal{W}}$.

\subsection{Sensor Data Preprocessing}
\label{sec:preprocessing}
The inputs to DLIOM are a dense 3D point cloud collected by a modern 360$^\circ$ mechanical LiDAR, such as an Ouster or a Velodyne (10-20Hz), in addition to time-synchronized linear acceleration and angular velocity measurements from a 6-axis IMU at a much higher rate (100-500Hz). To minimize information loss, we do not preprocess the point cloud except for a box filter of size 1m$^3$ around the origin which removes points that may be from the robot itself, and a light voxel filter for higher resolution clouds. This distinguishes our work from others that either attempt to detect features (e.g., corners, edges, and/or surfels) or aggressively downsamples the input cloud. On average, the point clouds used in this work on our custom platform contained ${\sim}$16,000 points per scan.

In addition, prior to downstream tasks, all sensor data is transformed to be in $\mathcal{R}$ located at the robot's center of gravity via extrinsic calibration. For LiDAR, each acquired scan is rotated and shifted via $\prescript{\mathcal{R}}{\mathcal{L}}{\textbf{T}} \in \mathbb{SE}(3)$ such that $[\,p^\mathcal{R} \,\, 1\,]^\top = \prescript{\mathcal{R}}{\mathcal{L}}{\textbf{T}} \, [\,p^\mathcal{L} \,\, 1\,]^\top$ for each point in the scan. For IMU, effects of displacing linear acceleration measurements on a rigid body must be considered if the sensor is not located exactly at the center of gravity. This is compensated for by considering all contributions of linear acceleration at $\mathcal{R}$ via the cross product of angular velocity with the displacement between the IMU and center of gravity, such that for raw linear acceleration $\boldsymbol{a}^\mathcal{B}_i$ measured in the IMU's frame, the corresponding linear acceleration in frame $\mathcal{R}$ is, assuming a constant displacement,
\begin{equation}
    \hat{\boldsymbol{a}}^\mathcal{R}_i = \hat{\boldsymbol{a}}^\mathcal{B}_i + \left[ \dot{\hat{\boldsymbol{\omega}}}^\mathcal{R}_i \times \prescript{\mathcal{R}}{\mathcal{B}}{\textbf{t}} \right] + \left[ \hat{\boldsymbol{\omega}}^\mathcal{R}_i \times \left( \hat{\boldsymbol{\omega}}^\mathcal{R}_i \times \prescript{\mathcal{R}}{\mathcal{B}}{\textbf{t}} \right) \right] \\
\end{equation}
\noindent where $\prescript{\mathcal{R}}{\mathcal{B}}{\textbf{t}}$ is the translational distance from $\mathcal{B}$ to $\mathcal{R}$, and $\hat{\boldsymbol{\omega}}^\mathcal{R}_i$ is $\hat{\boldsymbol{\omega}}^\mathcal{B}_i$ but only rotated to be in axis convention since angular velocity is equivalent at all points on a rigid body.

\subsection{Continuous-Time Motion Correction with Integrated Prior}

\begin{figure}[!t]
    \centering
    \includegraphics[width=0.95\columnwidth]{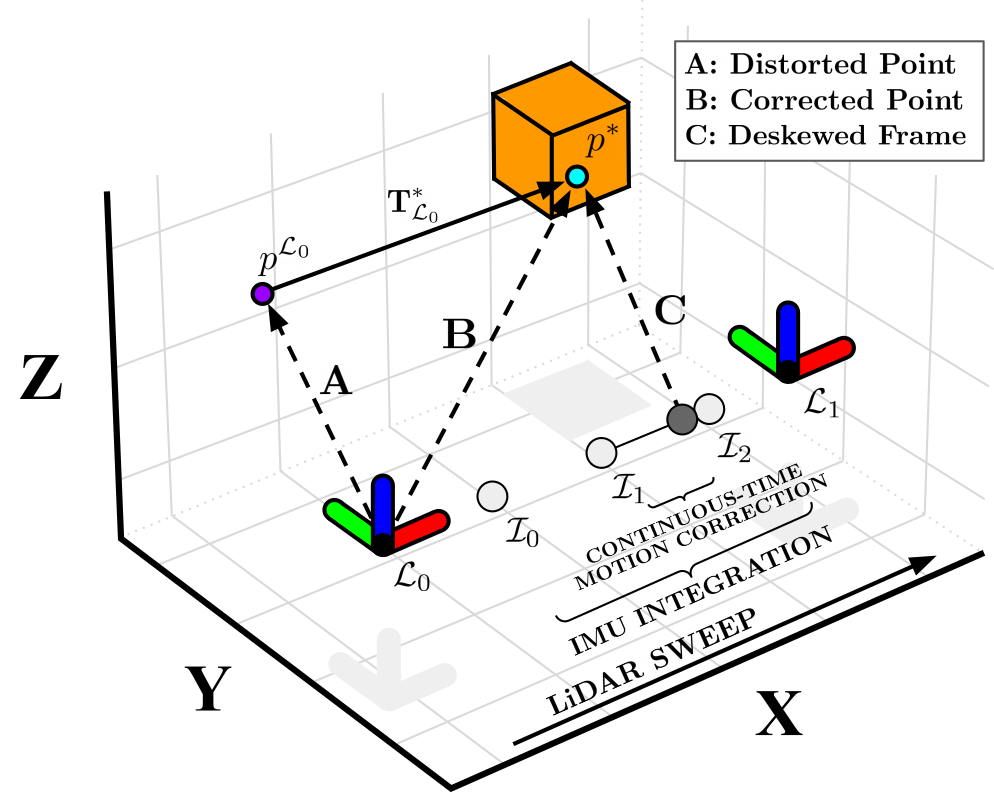}
    \caption{\textbf{Coarse-to-Fine Point Cloud Deskewing.} A distorted point $p^{\mathcal{L}_0}$ (A) is deskewed through a two-step process which first integrates IMU measurements between scans, then solves for a unique transform in continuous-time (C) for the original point which deskews $p^{\mathcal{L}_0}$ to $p^*$ (B).}
    \label{fig:motion_correction}
    \vskip -0.2in
\end{figure}

Point clouds from spinning LiDAR sensors suffer from motion distortion during movement due to the rotating laser array collecting points at different instances during a sweep. Rather than assuming simple motion (i.e., constant velocity) during sweep that may not accurately capture fine movement, we instead use a more accurate constant jerk and angular acceleration model to compute a unique transform for each point via a two-step coarse-to-fine propagation scheme. This strategy aims to minimize the errors that arise due to the sampling rate of the IMU and the time offset between IMU and LiDAR point measurements. 
Trajectory throughout a sweep is first coarsely constructed through numerical IMU integration, which is subsequently refined by solving a set of analytical continuous-time equations in $\mathcal{W}$ (Fig.~\ref{fig:motion_correction}).

\begin{figure}[!t]
    \centering
    \includegraphics[width=0.95\columnwidth]{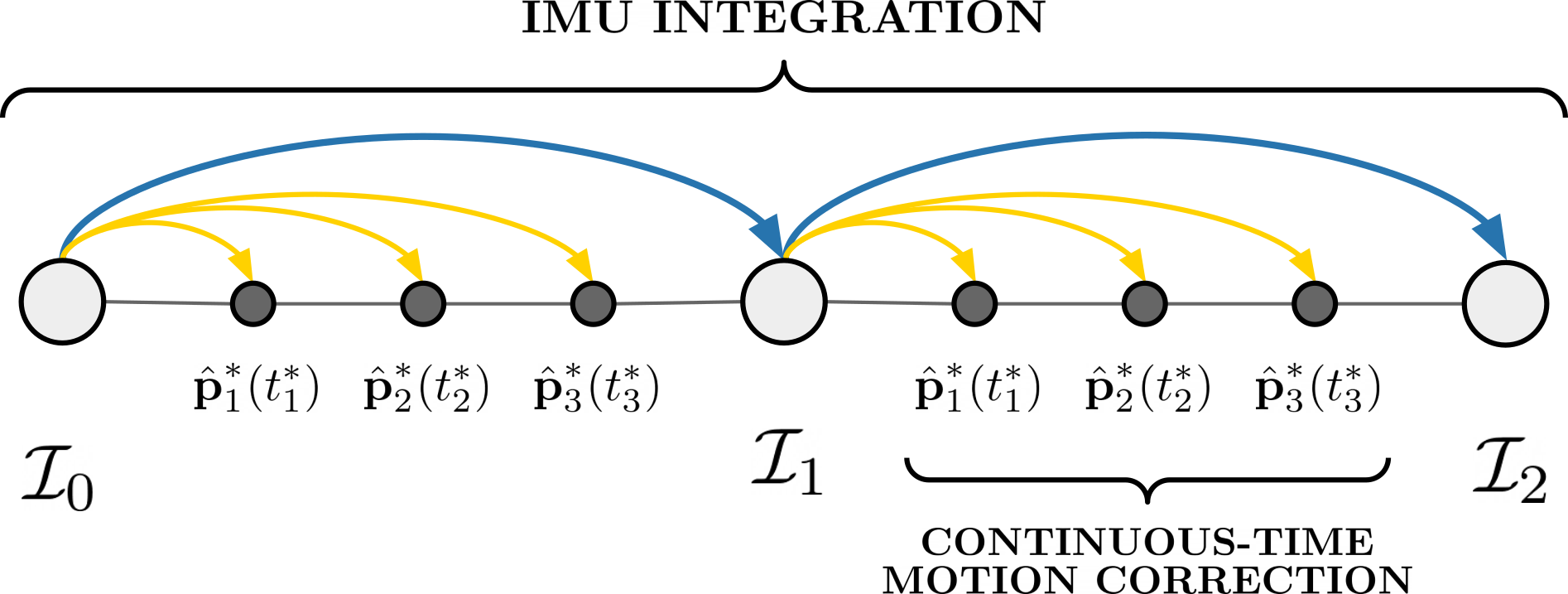}
    \caption{\textbf{Continuous-Time Motion Correction.} For each point in a cloud, a unique transform is computed by solving a set of closed-form motion equations parameterized solely by its timestamp to provide accurate, point-wise continuous-time motion correction.}
    \label{fig:continuous_time}
    \vskip -0.2in
\end{figure}

Let $t_{k}$ be the clock time of the received point cloud $\mathcal{P}_k^{\mathcal{R}}$ with $N$ number of points, and let $t_{k} + \Delta t_{k}^n$ be the timestamp of a point $p_{k}^n$ in the cloud. To approximate each point's location in $\mathcal{W}$, we first integrate IMU measurements between $t_{k\text{-}1}$ and $t_{k} + \Delta t_{k}^N$ via
\begin{equation}
\begin{alignedat}{2}
    \label{eq:deskew}
    \hat{\textbf{p}}_{i} &= \hat{\textbf{p}}_{i\text{-}1} &&+ \hat{\textbf{v}}_{i\text{-}1}\Delta t_i + \tfrac{1}{2}\hat{\textbf{R}}(\hat{\textbf{q}}_{i\text{-}1})\hat{\boldsymbol{a}}_{i\text{-}1}\Delta t_i^2 + \tfrac{1}{6}\hat{\boldsymbol{j}}_i\Delta t_i^3 \,,\\
    \hat{\textbf{v}}_{i} &= \hat{\textbf{v}}_{i\text{-}1} &&+ \hat{\textbf{R}}(\hat{\textbf{q}}_{i\text{-}1})\hat{\boldsymbol{a}}_{i\text{-}1}\Delta t_i \,,\\
    \hat{\textbf{q}}_{i} &= \hat{\textbf{q}}_{i\text{-}1} &&+ \tfrac{1}{2} (\hat{\textbf{q}}_{i\text{-}1} \otimes \hat{\boldsymbol{\omega}}_{i\text{-}1} )\Delta t_i + \tfrac{1}{4}(\hat{\textbf{q}}_{i\text{-}1} \otimes \hat{\boldsymbol{\alpha}_i}) \Delta t_i^2\,,
\end{alignedat}
\end{equation}
\noindent for $i = 1,\hdots,M$ for $M$ number of IMU measurements between two scans, where $\hat{\boldsymbol{j}}_i = \tfrac{1}{\Delta t_i} {(\hat{\textbf{R}}(\hat{\textbf{q}}_{i})\hat{\boldsymbol{a}}_i - \hat{\textbf{R}}(\hat{\textbf{q}}_{i\text{-}1})\hat{\boldsymbol{a}}_{i\text{-}1})}$ and $\hat{\boldsymbol{\alpha}}_i =  \tfrac{1}{\Delta t_i} {(\hat{\boldsymbol{\omega}}_i - \hat{\boldsymbol{\omega}}_{i\text{-}1})}$ are the estimated linear jerk and angular acceleration, respectively. The set of homogeneous transformations $\hat{\textbf{T}}_i^{\mathcal{W}} \in \mathbb{SE}(3)$ that correspond to $\hat{\textbf{p}}_i$ and $\hat{\textbf{q}}_i$ then define the coarse, \textit{discrete}-time trajectory during a sweep. Then, an analytical, \textit{continuous}-time solution from the nearest preceding transformation to each point $p_k^n$ recovers the point-specific deskewing transform $\hat{\textbf{T}}^{\mathcal{W}*}_{n}$, such that
\begin{equation}
\begin{alignedat}{2}
    \label{eq:deskew_timestamp}
        \hat{\textbf{p}}^*(t) &= \hat{\textbf{p}}_{i\text{-}1} &&+ \hat{\textbf{v}}_{i\text{-}1} t + \tfrac{1}{2}\hat{\textbf{R}}(\hat{\textbf{q}}_{i\text{-}1})\hat{\boldsymbol{a}}_{i\text{-}1} t^2 + \tfrac{1}{6}\hat{\boldsymbol{j}}_i t^3 \,,\\
        \hat{\textbf{q}}^*(t) &= \hat{\textbf{q}}_{i\text{-}1} &&+ \tfrac{1}{2} (\hat{\textbf{q}}_{i\text{-}1} \otimes \hat{\boldsymbol{\omega}}_{i\text{-}1}) t + \tfrac{1}{4} (\hat{\textbf{q}}_{i\text{-}1} \otimes \hat{\boldsymbol{\alpha}_i}) t^2 \,,
\end{alignedat}
\end{equation}
\noindent where ${i{-}1}$ and $i$ correspond to the closest preceding and successive IMU measurements, respectively, $t$ is the timestamp between point $p_k^n$ and the closest preceding IMU, and $\hat{\textbf{T}}^{\mathcal{W}*}_{n}$ is the transformation corresponding to $\hat{\textbf{p}}^*$ and $\hat{\textbf{q}}^*$ for $p_k^n$ (Fig.~\ref{fig:continuous_time}). Note that (\ref{eq:deskew_timestamp}) is parameterized only by $t$ and therefore a transform can be queried for any desired time to construct a continuous-time trajectory.

The result of this two-step procedure is a motion-corrected point cloud that is also approximately aligned with the map in $\mathcal{W}$, which therefore inherently incorporates the optimization prior used for GICP (Sec.~\ref{sec:method:s2m}). Importantly, (\ref{eq:deskew}) and (\ref{eq:deskew_timestamp}) depend on the accuracy of $\hat{\textbf{v}}^{\mathcal{W}}_0$, the initial estimate of velocity, $\textbf{b}_k^a$ and $\textbf{b}_k^\omega$, the estimated IMU biases, in addition to an accurate initial body orientation $\hat{\textbf{q}}_0$ (to properly compensate for the gravity vector) at the time of motion correction. We therefore emphasize that, a key to the reliability of our approach is the \textit{guaranteed global convergence} of these terms by leveraging a nonlinear geometric observer \cite{lopez2023contracting}, provided that scan-matching returns an accurate solution.

\section{Robust \& Perceptive Localization}

\subsection{Slip-Resistant Keyframing via Sensor-Agnostic Degeneracy}

\begin{figure}[!t]
    \centering
    \includegraphics[width=0.95\columnwidth]{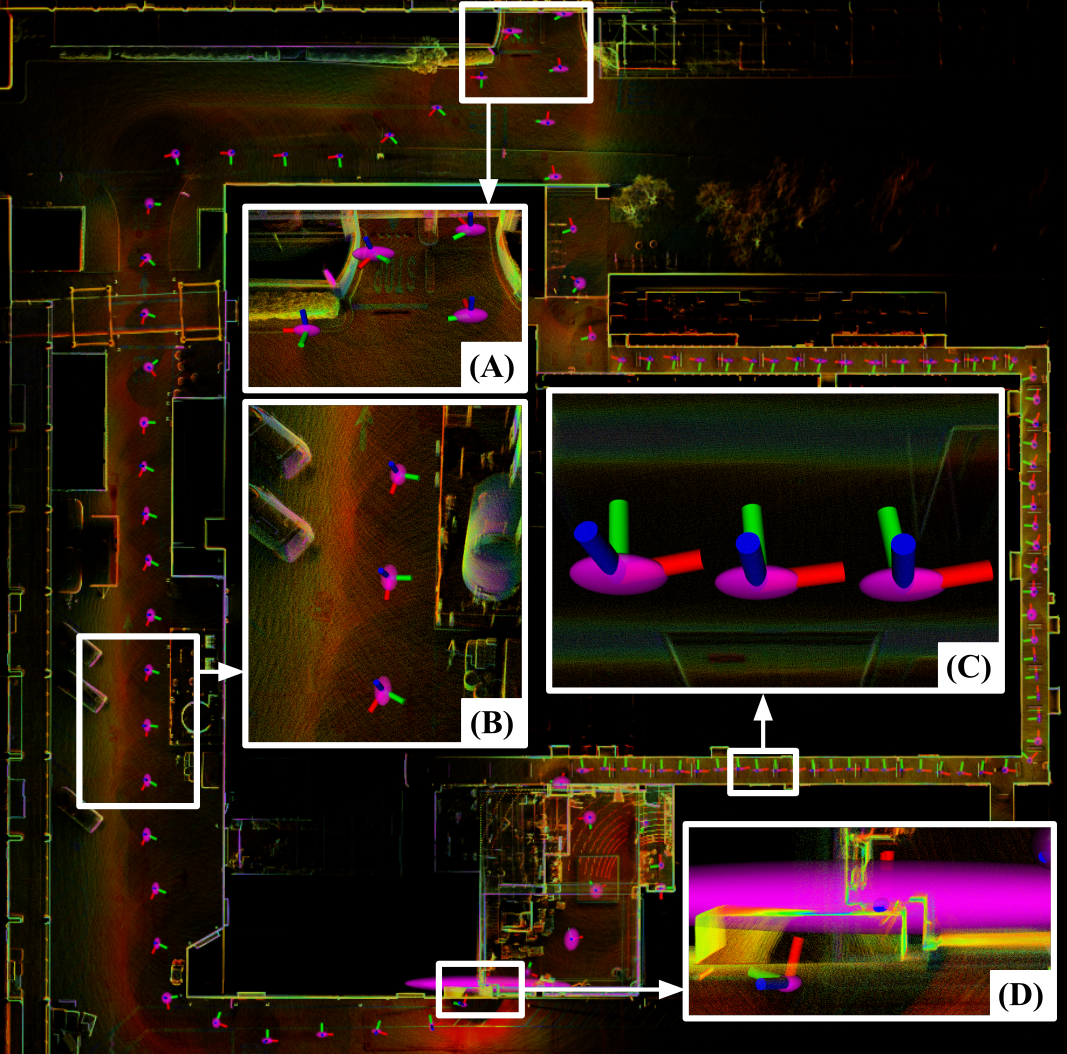}
    \caption{\textbf{Sensor-Agnostic Degeneracy.} Uncertainty ellipsoids (purple) for each keyframe computing using our generalized degeneracy metric in (A \& B) outdoor environments, (C) a narrow hallway, and (D) through a doorway. Our metric is global in that the ellipsoids are consistent in size in both indoor and outdoor environments; our metric is also sensor-agnostic in that it accounts for the density of the cloud (which can vary across different LiDAR sensors and voxelization leaf sizes). Note that these ellipsoids usually on the millimeter-scale but have been enlarged for visualization clarity.}
    \label{fig:degeneracy}
    \vskip -0.2in
\end{figure}

Convergence of (\ref{eq:gicp}) into a sub-optimal local minima can occur when correspondences for GICP plane-to-plane registration are sparse or insufficient. Such weak data correspondences can subsequently lead to poor or diverging localization due to the estimate ``escaping" a shallow gradient around the local minima. This phenomenon, often referred to as \textit{LiDAR slippage}, often occurs when the surrounding environment is featureless or otherwise geometrically degenerate (e.g., long tunnels or large fields) and is a result of incorrect data association between the source and target point clouds. Ill-constrained optimization problems can also arise in \textit{keyframe-based} LIO when the extracted submap insufficiently represents the surrounding environment and which results in a low number of data correspondences for scan-to-map matching. This can happen when there is an abrupt change in the environment (e.g., walking through a door or up a stairwell) but there are no nearby keyframes which describe the new environment. While previous works have used the condition number of the Hessian \cite{tagliabue2020lion, ebadi2021dare, jiao2023dams} to identify environmental degeneracy, such that $\kappa(\mathbf{H_{tt}}) = {|\lambda_\text{max}(\mathbf{H_{tt}})|}\,/\,{|\lambda_\text{min}(\mathbf{H_{tt}})|}$, this metric informs a system only of \textit{relative} slippage with respect to the most constrained and least constrained directions of the problem. The condition number is affected by the size of the environment and the density of points, and therefore a more robust approach is to compute a more consistent metric of degeneracy across differently-sized environments and sensor configurations. This enables a global, sensor-agnostic metric in which we use to detect when a new keyframe should be inserted into the environment.

Let $\mathcal{C}$ be the set of all corresponding points between $\hat{\mathcal{P}}_{k}^{\mathcal{W}}$ and $\hat{\mathcal{S}}_{k}^{\mathcal{W}}$, and therefore $|\mathcal{C}|$ is the total number of corresponding points, and let $\mathcal{E}$ be the total error between all correspondences after convergence of a nonlinear least squares solver (such as Levenberg-Marquardt) as described previously in (\ref{eq:gicp}). Also, let $\mathbf{H} \in \mathbb{R}^{6x6}$ be the Hessian of GICP, and let $\mathbf{H_{tt}} \in \mathbb{R}^{3x3}$ be the submatrix corresponding to the translational portion of $\mathbf{H}$, with eigenvalues $\lambda_{\text{max}}(\mathbf{H_{tt}}) \geq \cdots \geq \lambda_{\text{min}}(\mathbf{H_{tt}})$ which provide information regarding the local gradient of the nonlinear optimization after convergence. Note that $\mathbf{H} \approx \mathbf{J}^\top \mathbf{J}$ for computational efficiency and $\mathbf{J}$ is the Jacobian. Then, the \textit{global degeneracy} $d_k$ of the system is the maximum value after scaling each of the eigenvalues $\lambda(\mathbf{H_{tt}})$, such that
\begin{equation}
    \label{eq:degeneracy}
    d_k = \text{max} \left[ \, \frac{ m_k^2 }{ \lambda(\mathbf{H_{tt}}) \,\, \sqrt{z_k}} \, \right] \,,
\end{equation}
\noindent where $m_k$ is the computed \textit{spaciousness} \cite{chen2022direct}, defined as $m_k = \alpha m_{k-1} + \beta M_k$, where $M_k$ is the median Euclidean point distance from the origin to each point in the preprocessed point cloud (with constants $\alpha$ = $0.95$, $\beta$ = $0.05$), and $z_k$ is the cloud \textit{sparsity} as defined above in (\ref{eq:sparsity}). This degeneracy is computed for each incoming scan and saved in-memory for each new keyframe. If the difference between the current degeneracy and degeneracy at the location of the previous keyframe is sufficiently large, a new keyframe is inserted to provide the scan-to-map module with new information. 

\begin{figure}[!t]
    \centering
    \includegraphics[width=0.95\columnwidth]{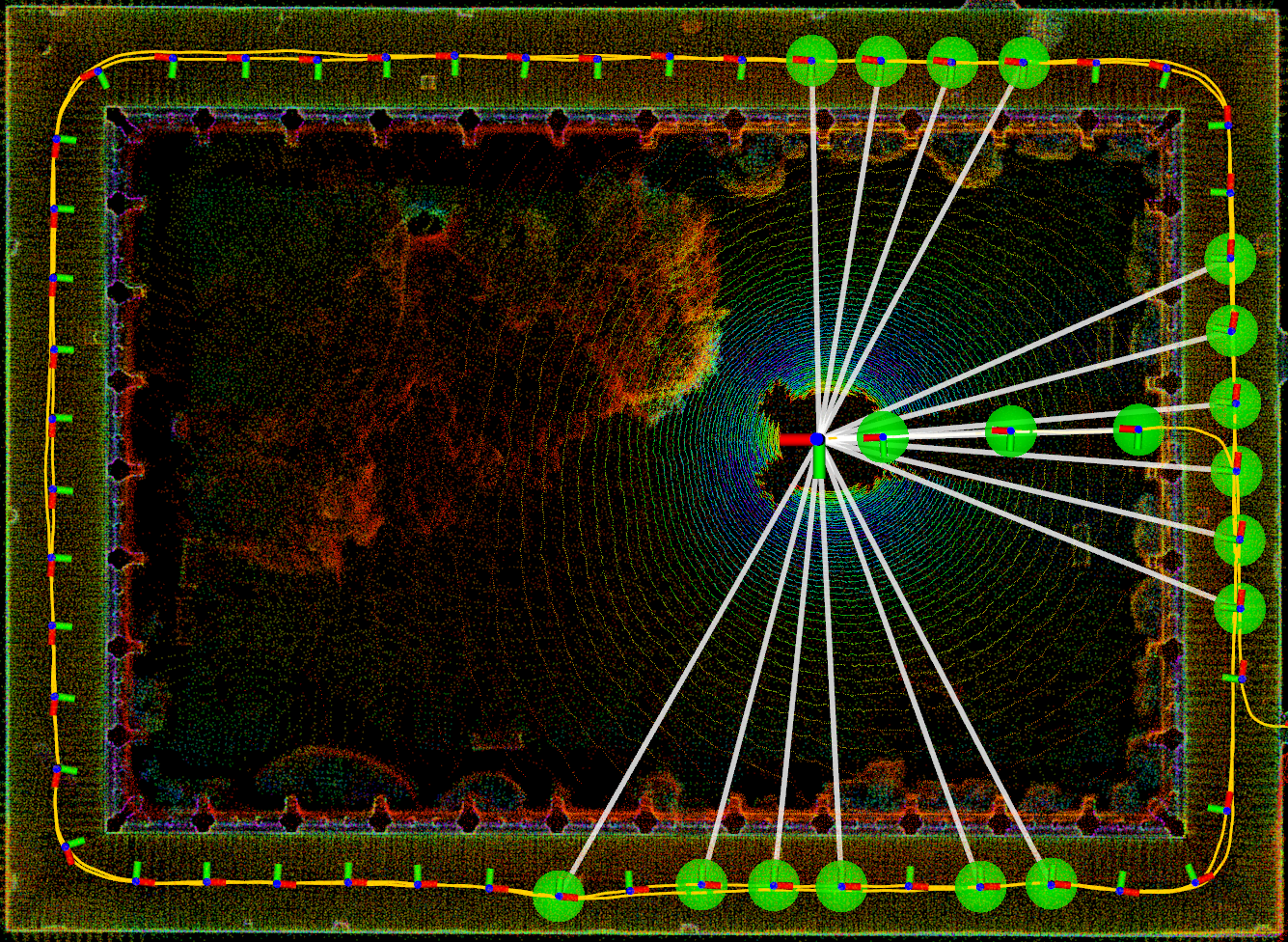}
    \caption{\textbf{Submapping via Jaccard Index.} Submap generation for the scan-to-map stage using the Newer College Dataset Extension - Cloister in Collection 2 \cite{zhang2021multicamera}. For each newly acquired scan, we compute its Jaccard index against each environmental keyframe (axes) and extract only those which have a significant overlap with the current scan (green circles \& white lines). The point clouds associated to the overlapping keyframes are then concatenated, alongside their in-memory covariances, for accurate scan-to-map registration. A threshold of at least $20\%$ overlap was used in this example.}
    \label{fig:submap_jaccard}
    \vskip -0.2in
\end{figure}

The intuition behind (\ref{eq:degeneracy}) lies in how each scaling factor (i.e., $m_k$ and $z_k$) affects $\lambda(\mathbf{H_{tt}})$. In particular, while computing the condition number $\kappa(\mathbf{H_{tt}})$ can provide an idea of how ellipsoidal the local gradient is (and therefore how long it may take to converge to a local minimum), an elongated gradient does not necessarily indicate the onset of slippage from being poorly constrained. In other words, $\kappa(\mathbf{H_{tt}})$ is a \textit{relative} metric of how well the optimization problem is constrained, since it only computes the relative ratio between the steepest and shallowest directions. To get a more accurate idea of when slippage may occur, (\ref{eq:degeneracy}) directly looks at (the inverse of) each individual eigenvalue. By rewarding sensors which provide less information about the environment via $z_k$, and by penalizing larger environments since measurements are less accurate with increasing distance, these various scaling factors allow $d_k$ to be more consistent across different sensors and differently sized environments, which enables a more reliable metric of slip-detection (Fig.~\ref{fig:degeneracy}).

\subsection{Submap Generation via 3D Jaccard Index}
\label{sec:jaccard}

A key innovation of the DLIOM algorithm is how it explicitly derives its keyframe-based submap for scan-to-map registration.
Ideally, the full history of all observed points would be matched against to ensure that there is no absence of important environmental information during scan-matching.
Unfortunately, this is far too computationally intractable due to the sheer number of nearest-neighbor operations required for aligning against such a large map.
Whereas previous approaches either naively assume that the closest points in map-space are those which are most relevant, or they implicitly compute keyframe relevancy via nearest neighbor and convex hull extraction in keyframe-space \cite{chen2022direct}, we propose a new method for deriving the local submap that explicitly maximizes coverage between the current scan and the submap by computing the $\textit{Jaccard index}$ \cite{jaccard1912distribution} between the current scan and each keyframe.

Let the intersection between two point clouds $\mathcal{P}_1 \cap \mathcal{P}_2$ be a set $\mathcal{C}_{1,2}$ which contains all corresponding points between the two clouds in a common reference frame (within some corresponding distance), and therefore let $|\mathcal{C}_{1,2}|$ be the total number of corresponding points. In addition, let the union between two point clouds $\mathcal{P}_1 \cup \mathcal{P}_2$ be defined as the set $\mathcal{U}_{1,2}$ which contains all non-intersecting points between the two point clouds, in addition to the mean of each pair of corresponding points in $\mathcal{C}_{1,2}$, such that the total number of points in $\mathcal{U}_{1,2}$ equates to
\begin{equation}
    |\mathcal{U}_{1,2}| = \left( |\mathcal{P}_1| \, \oplus |\mathcal{P}_2| \, \right) \, \setminus \, |\mathcal{C}_{1,2}| \,.
\end{equation}
Then, for each newly acquired LiDAR scan at time $k$, we compute the 3D Jaccard index between the scan $\hat{\mathcal{P}}_k^{\mathcal{W}}$ and each $j^{\text{th}}$ keyframe $\mathcal{K}_j^{\mathcal{W}}$, defined as
\begin{equation}
    J(\hat{\mathcal{P}}_k^{\mathcal{W}}, \mathcal{K}_j^{\mathcal{W}})  = \frac{|\mathcal{\hat{\mathcal{P}}}_k^{\mathcal{W}} \cap \mathcal{K}_j^{\mathcal{W}}|}{|\mathcal{\hat{\mathcal{P}}}_k^{\mathcal{W}} \cup \mathcal{K}_j^{\mathcal{W}}|} \,,
    \label{eq:jaccard}
\end{equation}
\noindent or, in otherwords, the equivalent of the ``intersection over union" similarity measurement in the 3D domain. If $J(\hat{\mathcal{P}}_k^{\mathcal{W}}, \mathcal{K}_j^{\mathcal{W}})$ surpasses a set threshold for the $j^{\text{th}}$ keyframe (i.e., a keyframe is sufficiently similar), then that keyframe is included within the submap to be used for scan-to-map registration. In contrast to previous methods which derive the submap through a series of heuristics (such as directly retrieving local points within a certain radius of the current position or assuming that nearby keyframes contain relevant points) our method explicitly computes each keyframes' relevancy to the current environment to ensure the scan-to-map optimization is well-constrained with maximum coverage between the scan and the submap. In addition, by using only keyframe scans that contain significant overlap with the current scan, this guarantees that there are no wasted operations when building normals or the kdtree data structure for the submap (Fig.~\ref{fig:submap_jaccard}). 

\subsection{Adaptive Scan-Matching via Cloud Sparsity}
\label{sec:method:s2m}

\begin{figure}[!t]
    \centering
    \includegraphics[width=0.95\columnwidth]{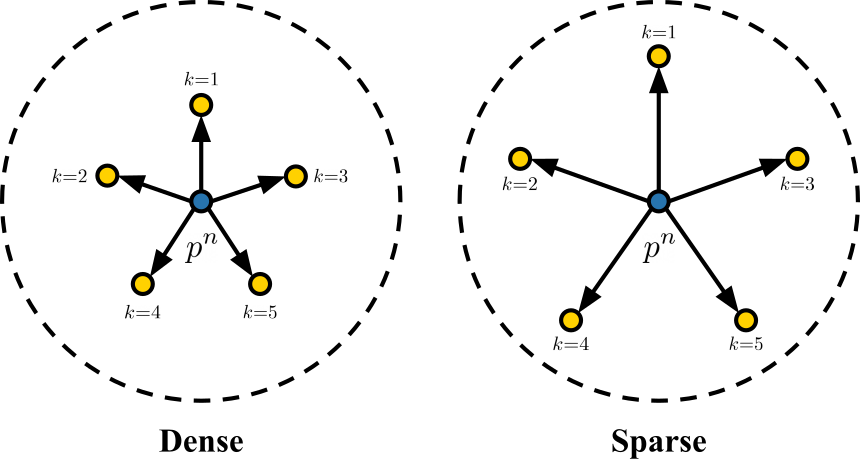}
    \caption{\textbf{Adaptive Scan-Matching via Cloud Sparsity.} For each motion-corrected point cloud, we compute its \textit{sparsity}, defined as the average per-point Euclidean distance across $K$ nearest neighbors (\ref{eq:sparsity}) ($K{=}5$ in this example). This metric is used to scale the scan-to-map module's maximum correspondence distance for adaptive registration. A scan within a small-scale environment will contain points much closer together (left), so a small movement will have a small effect on point displacement. On the otherhand, a large environment will have points much more spread out (right) and will require a larger search distance during GICP for correct data association.}
    \label{fig:sparsity}
    \vskip -0.2in
\end{figure}

By simultaneously correcting for motion distortion and incorporating the GICP optimization prior into the point cloud, DLIOM can directly perform scan-to-map registration and bypass scan-to-scan required in previous methods. This registration is cast as a nonlinear optimization problem which minimizes the distance of corresponding points/planes between the current scan and an extracted local submap. Let $\hat{\mathcal{P}}_k^{\mathcal{W}}$ be the corrected cloud in $\mathcal{W}$ and $\hat{\mathcal{S}}_k^{\mathcal{W}}$ be the extracted submap. Then, the objective of scan-to-map optimization is to find a transformation $\Delta \hat{\textbf{T}}_k$ which better aligns the point cloud, where
\begin{equation}
    \label{eq:gicp}
    \Delta \hat{\textbf{T}}_k = \argmin_{\Delta \textbf{T}_k} \, \mathcal{E} \left( \Delta \textbf{T}_k \hat{\mathcal{P}}_{k}^{\mathcal{W}},\, \hat{\mathcal{S}}_{k}^{\mathcal{W}} \right) \,,
\end{equation}
\noindent such that the GICP residual error $\mathcal{E}$ is defined as
\begin{equation*}
    \mathcal{E} \left( \Delta \textbf{T}_k \hat{\mathcal{P}}_{k}^{\mathcal{W}}, \hat{\mathcal{S}}_{k}^{\mathcal{W}} \right) = \sum_{c \in \mathcal{C}} d_c^\top \left( C_{k,c}^{\mathcal{S}} + \Delta \textbf{T}_k C_{k,c}^{\mathcal{P}} \Delta \textbf{T}^\top_k \right)^{-1} d_c \,,
\end{equation*}
\noindent for a set $\mathcal{C}$ of corresponding points between $\hat{\mathcal{P}}_k^{\mathcal{W}}$ and $\hat{\mathcal{S}}_k^{\mathcal{W}}$ at timestep $k$,\, $d_c = \hat{s}_k^c - \Delta \textbf{T}_k \hat{p}_k^c$,\, $\hat{p}_k^c \in \hat{\mathcal{P}}_{k}^{\mathcal{W}}$,\, $\hat{s}_k^c \in \hat{\mathcal{S}}_{k}^{\mathcal{W}}$,\, $\forall c \in \mathcal{C}$, and $C_{k,c}^{\mathcal{P}}$ and $C_{k,c}^{\mathcal{S}}$ are the estimated covariance matrices for point cloud $\hat{\mathcal{P}}_k^{\mathcal{W}}$ and submap $\hat{\mathcal{S}}_k^{\mathcal{W}}$, respectively. Then, following \cite{segal2009generalized}, this point-to-plane formulation is converted into a plane-to-plane optimization by regularizing covariance matrices $C_{k,c}^{\mathcal{P}}$ and $C_{k,c}^{\mathcal{S}}$ with $(1, 1, \epsilon)$ eigenvalues, where $\epsilon$ represents the low uncertainty in the surface normal direction. 
The resulting $\Delta \hat{\textbf{T}}_k$ represents an optimal correction transform which better globally aligns the prior-transformed scan $\hat{\mathcal{P}}_k^{\mathcal{W}}$ to the submap $\hat{\mathcal{S}}_k^{\mathcal{W}}$, so that $\hat{\textbf{T}}^{\mathcal{W}}_k = \Delta \hat{\textbf{T}}_k \hat{\textbf{T}}_M^\mathcal{W}$ (where $\hat{\textbf{T}}_M^\mathcal{W}$ is the last point's IMU integration) is the globally-refined robot pose which is used for map construction and as the update signal for the nonlinear geometric observer.

An important parameter that is often overlooked is the maximum distance at which corresponding points or planes should be considered in the optimization. This parameter is often hand-tuned by the user but should scale with the environmental structure for consistency and computational efficiency. For example, in small-scale environments (e.g., a lab room), points in the LiDAR scan are much closer together so a small movement has a small effect on the displacement of a given point. In contrast, the same point in a more open environment (e.g., a point on a distant tree outside) will be displaced farther with a small rotational movement due to a larger distance and therefore needs a greater search radius for correct correspondence matching (Fig.~\ref{fig:sparsity}). Thus, we set the GICP solver's maximum correspondence search distance between two point clouds according to the ``sparsity" of the current scan, defined as $z_k = \alpha z_{k-1} + \beta D_k$, where
\begin{equation}
    D_k = \frac{1}{|\mathcal{P}| N} \sum_{n=1}^{N} D^n_k
    \label{eq:sparsity}
\end{equation}
is the normalized per-point sparsity, $D^n_k$ is the average Euclidean distance to $K$ nearest neighbors for point $n$, and $\alpha=0.95$ and $\beta=0.05$ are smoothing constants to produce $z_k$, the filtered signal set as the max correspondence distance. Intuitively, this is the average inter-point distance in the current scan; the larger the environment, higher the number of sparse points (i.e., points further away), driving this number up. By adapting the corresponding distance according to the sparsity of points, the efficacy of scan-matching can be more consistent across differently-sized environments.

\subsection{Hierarchical Geometric Observer}
\label{sec:methods:geo}

The transformation $\hat{\textbf{T}}^{\mathcal{W}}_k$ computed by scan-to-map alignment is fused with IMU measurements to generate a full state estimate $\hat{\textbf{X}}_k$ via a novel hierarchical nonlinear geometric observer.  
A full analysis of the observer can be found in \cite{lopez2023contracting}, but in summary, one can show that $\hat{\textbf{X}}$ will globally converge to $\textbf{X}$ in the deterministic setting with minimal computation.
The proof utilizes contraction theory to first prove that the quaternion estimate converges exponentially to a region near the true quaternion.
The orientation estimate then serves as an input to another contracting observer that estimates translation states.
This architecture forms a contracting hierarchy that guarantees the estimates converge to their true values.
The strong convergence property of the observer is the main advantage over other fusion schemes, e.g., filtering or pose graph optimization that possess minimal convergence guarantees, and will be important for future theoretical studies on the advantages of our LiDAR odometry and mapping pipeline.
From a practical viewpoint, the observer generates smooth estimates in real-time so it output is also suitable for  control.
The observer used here is a special case of that in \cite{lopez2023contracting}.


\begin{figure*}[!t]
    \centering
    \includegraphics[width=0.99\textwidth]{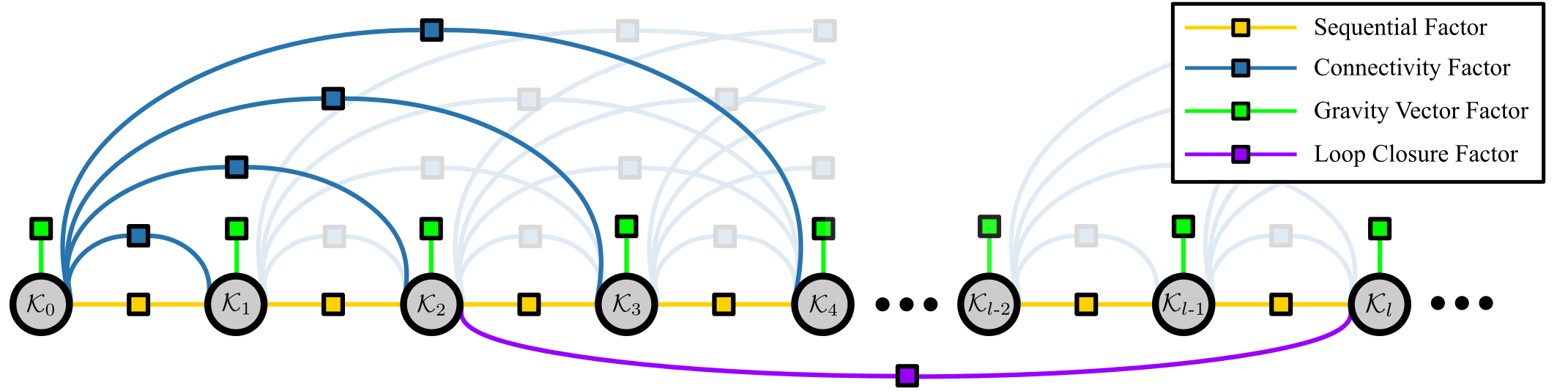}
    \caption{\textbf{Keyframe-based Factor Graph Mapping.} Our mapper adds a node to its factor graph for each new keyframe and adds relative constraints through either sequential factors (yellow), connectivity factors (blue), gravity factors (green), or loop closure factors (purple). Sequential factors provide a strong ``skeleton" for the graph with low uncertainty between adjacent keyframes, while connectivity factors scale depending on the overlap between pairs of keyframes. Loop closure factors enable global consistency after long-term drift from pure odometry.}
    \label{fig:factorgraph}
    \vskip -0.2in
\end{figure*}

Let $\gamma_{\ell\in\{1,\dots,5\}}$ be positive constants and $\Delta t^{+}_k$ be the time between GICP poses.
If the errors between propagated and measured poses are $\textbf{q}_e \coloneqq (q_e^\circ,~\vec{q}_e) = \hat{\textbf{q}}^*_i \otimes \hat{\textbf{q}}_k$ and $\textbf{p}_e = \hat{\textbf{p}}_k - \hat{\textbf{p}}_i$, then the attitude update takes the form of
\begin{equation}
    \begin{alignedat}{3}
    \label{eq:update_attitude}
      &\hat{\textbf{q}}_i && \leftarrow \hat{\textbf{q}}_i &&+ \Delta t^{+}_k \, \gamma_1 \, \hat{\textbf{q}}_i \otimes \left[ \begin{array}{c} 1 - |q_e^\circ| \\ \mathrm{sgn}(q_e^\circ) \, \vec{q}_e \end{array} \right] \,,\\
      &\hat{\textbf{b}}^{\boldsymbol{\omega}}_i && \leftarrow \hat{\textbf{b}}^{\boldsymbol{\omega}}_i &&- \Delta t^{+}_k \, \gamma_2 \, q_e^\circ \vec{q}_e  \,,\\
    \end{alignedat}
\end{equation}
and the translational update as
\begin{equation}
    \begin{alignedat}{3}
    \label{eq:update_translation}
      &\hat{\textbf{p}}_i && \leftarrow \hat{\textbf{p}}_i &&+ \Delta t^{+}_k \, \gamma_3 \, \textbf{p}_e  \,,\\
      &\hat{\textbf{v}}_i && \leftarrow \hat{\textbf{v}}_i &&+ \Delta t^{+}_k \, \gamma_4 \, \textbf{p}_e  \,,\\
      &\hat{\textbf{b}}^{\boldsymbol{a}}_i && \leftarrow \hat{\textbf{b}}^{\boldsymbol{a}}_i &&- \Delta t^{+}_k \, \gamma_5 \, \hat{\textbf{R}}(\hat{\textbf{q}}_i)^\top \textbf{p}_e \,.\\
    \end{alignedat}
\end{equation}
Note that state correction is hierarchical as the attitude update (\ref{eq:update_attitude}) is completely decoupled from the translation update (\ref{eq:update_translation}).
Also, this is a fully nonlinear update which allows one to guarantee the state estimates are accurate enough to directly perform scan-to-map registration solely with an IMU prior without the need for scan-to-scan.

\section{Connective Mapping}

\begin{figure*}[!ht]
    \centering
    \includegraphics[width=0.99\textwidth]{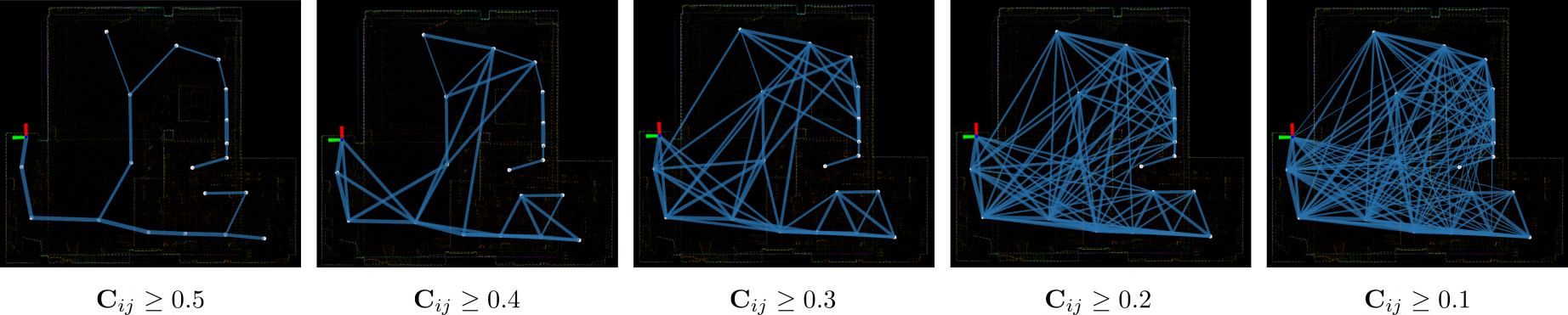}
    \caption{\textbf{Environmental Connectivity.} Example of increasing graph strength (left to right) by reducing the threshold for connective factors. A weak graph (left) is less locally accurate but allows for more compliancy when adding loop closures to the graph, while a strong graph (right) is more locally accurate from its higher number of interkeyframe factors, which are computed according to keyframe-to-keyframe overlap.}
    \label{fig:connectivity}
    \vskip -0.2in
\end{figure*}

Factor graphs are widely used in SLAM as they are a powerful tool to estimate a system's full state by combining pose estimates from various modalities via \textit{pose graph optimization} \cite{shan2020lio}. Such works model relative pose constraints as a maximum a posteriori (MAP) estimation problem with a Gaussian noise assumption, and they typically view mapping as an afterthought as a result of refining the trajectory. However, such a unimodal noise model is far too simplistic for the complex uncertainty distribution that can arise from LiDAR scan-matching and IMU pre-integration. Moreover, graph-based optimization for odometry possesses minimal convergence guarantees and can often result in significant localization error and map deformation from inconsistent sensor fusion. To this end, we instead employ a factor graph not for odometry (which is instead handled by our geometric observer), but rather to explicitly represent the environment. A new node is added to the graph for every incoming keyframe (as determined by DLIOM's odometry thread), and various factors between nodes contribute to the global consistency of the map (Fig.~\ref{fig:factorgraph}). In addition to the factors detailed in this section, we additionally add a gravity factor to locally constrain the direction of each keyframe as described in \cite{nemiroff2023gravity}.


\subsection{Connective Keyframe Factors}
\label{sec:connectivity}

Relative constraints between nodes in a factor graph are typically added sequentially (i.e., factors are added between adjacent nodes \cite{shan2020lio}), but the relationship between non-adjacent keyframes can provide additional information to the graph which helps to create a more accurate and globally consistent map. These additional constraints are what we call \textit{connective} factors, which are determined by pairs of keyframes having sufficient overlap in $\mathcal{W}$ as computed in Sec.~\ref{sec:jaccard}. That is, for $K$ number of total keyframes, the connectivity between $K_i^\mathcal{W}$ and $K_j^\mathcal{W}$ is defined as the 3D Jaccard index (\ref{eq:jaccard}) and encoded in a symmetric matrix $\mathbf{C} \in \mathbb{R}^{K\text{x}K}$, such that
\begin{equation}
    \mathbf{C}_{ij} = 
    \begin{dcases}
        \frac{|\mathcal{\hat{\mathcal{K}}}_i^{\mathcal{W}} \cap \mathcal{K}_j^{\mathcal{W}}|}{|\mathcal{\hat{\mathcal{K}}}_i^{\mathcal{W}} \cup \mathcal{K}_j^{\mathcal{W}}|} & \text{for } i > j \\
        \frac{|\mathcal{\hat{\mathcal{K}}}_j^{\mathcal{W}} \cap \mathcal{K}_i^{\mathcal{W}}|}{|\mathcal{\hat{\mathcal{K}}}_j^{\mathcal{W}} \cup \mathcal{K}_i^{\mathcal{W}}|}  & \text{for } i < j \\
        1 & \text{for } i = j \\
    \end{dcases}
\end{equation}
\noindent where the diagonal contains all $1$'s by definition of (\ref{eq:jaccard}). A new factor is added between two keyframes if $\mathbf{C}_{ij}$ is above a set threshold, and the noise for this factor is computed by $\zeta(1 - \mathbf{C}_{ij})$, where $\zeta$ is a tunable scaling parameter that controls the strength of the environmental graph (Fig.~\ref{fig:connectivity}).

\subsection{Keyframe-Based Loop Closures}
\label{sec:loopclosures}

Ideally, place recognition modules would search across all seen data for loop closures and add corresponding graph factors accordingly. However, storing all historical scans is computationally infeasible, so scans are stored in-memory incrementally as \textit{keyframes}. In this lens, such keyframes can be understood as the subset of all historical point clouds which maximize information about the environment and contain data about the most significant locations. However, individual scans can be quite sparse (depending on the selected sensor) and may not contain enough points for data association for accurate detection via scan-matching. Therefore, rather than iterating through all keyframes individually or reusing the submap constructed from the frontend which is only optimal for odometry \cite{wang2022dliom}, we instead build an additional submap optimized for the backend mapper which consists of all candidate loop closure keyframes.

After adding the new keyframe to the factor graph and the associated connectivity constraints as described above, prior to optimization we search for and perform loop closure detection through a three-step process. First, we extract keyframes that are within some radius of the current position, in addition to those that contain some overlap with the current keyframe. The corresponding point clouds are then concatenated into a \textit{loop cloud} $\mathcal{L}_k^\mathcal{W}$ and transformed back into $\mathcal{R}$, and GICP scan-matching is performed between this and the new keyframe. If the fitness score between $\hat{\mathcal{K}}_k^\mathcal{R}$ and $\mathcal{L}_k^\mathcal{R}$ is sufficiently low (i.e., the average Euclidean error across all corresponding points is small), a new loop-closure factor is added between the current keyframe and the closest keyframe in $\mathcal{W}$ used to build the loop cloud. Crucially, the registration is primed with a prior equal to the distance between these two keyframes in $\mathcal{W}$ with a sufficiently large plane-to-plane search distance. This process is fast since we do not rebuild the covariances required for GICP, as individual keyframe normals are concatenated instead \cite{chen2022direct}. This idea of reconstructing a submap for loop closure detection can easily be extend to using other place recognition modules (e.g., \cite{gkim-2018-iros, gskim-2021-tro}) for further robustness.

\section{Algorithmic Implementation}

This section highlights three important implementation details of our system for small lightweight platforms: sensor time synchronization, resource management for consistent computational load, and velocity-consistent loop closures.

\subsection{Sensor Synchronization}
Time synchronization is a critical element in odometry algorithms which utilize sensors that have their own internal clock.
This is necessary as it permits time-based data association to temporally align IMU measurements and LiDAR scans.
There are three clock sources in DLIOM: one each for the LIDAR, IMU, and processing computer.
Hardware-based time synchronization---where the acquisition of a LiDAR scan is triggered from an external source---is not compatible with existing spinning LiDARs since starting and stopping the rotation assembly can lead to inconsistent data acquisition and timing. 
As a result, we developed a software-based approach that compensates for the offset between the LiDAR (IMU) clock and the processing computer clock.
When the first LiDAR (IMU) packet is received, the processing computer records its current clock time $\prescript{c}{}t_0$ and the time the measurement was taken on the sensor $\prescript{s}{}t_0$. Then, each subsequent $k^{\text{th}}$ measurement has a time $\prescript{c}{}t_k$ with respect to the processing computer clock given by $\prescript{c}{}t_k = \prescript{c}{}t_0 + (\prescript{s}{}t_k \,-\, \prescript{s}{}t_0)$, where $\prescript{s}{}t_k$ is the time the measurement was taken on the sensor.
This approach was found to work well in practice despite its inability to observe the transportation delay of sending the first measurement over the wire.
The satisfactory performance was attributed to using the elapsed \emph{sensor} time to compute the compensated measurement time since a sensor's clock is generally more accurate than that of the processing computer.

\subsection{Submap Multithreading}
Fast and consistent computation time is essential for ensuring that incoming LiDAR scans are not dropped, especially on resource-constrained platforms. To this end, DLIOM offloads work not immediately relevant to the current scan to a separate thread which minimally interferes with its parent thread as it handles further incoming scans. Thus, the main point cloud processing thread has lower, more consistent computation times. The secondary thread builds the local submap kdtree used for scan-matching and builds data structures corresponding to each keyframe which are needed by the submap. Speed of the submap building process is additionally increased by saving in-memory the computed kdtrees for each keyframe in order to quickly compute the Jaccard index of each keyframe, making that process negligibly different than an implicit nearest neighbor keyframe search. This thread can finish at any time without affecting DLIOM's ability to accept new LiDAR scans. Additionally, it periodically checks if the main thread is running so it can pause itself and free up resources that may be needed by the highly-parallelized scan-matching algorithm. Crucially, the submap changes at a much slower rate than the LiDAR sensor rate, and there is no strict deadline for when the new submap must be built. Therefore, the effect of this thread---which is to occasionally delay a new submap by a few scan iterations---has negligible impact on performance. 

\subsection{Velocity-Consistent Loop Closures}
Although our odometry module constructs a submap of relevant keyframes, these keyframes are pulled directly from the globally optimized map. Therefore, we must be careful with how the state and keyframes get updated upon loop closure detection. In particular, the estimated position within the map will jump instantaneously, but a \textit{continuous} trajectory would be beneficial for control and require fewer updates in the odometry module. We therefore allow the mapping module to perform this instantaneous update and maintain the robot pose in the \textit{map} frame, but establish an offset from the \textit{odometry} frame so that the robot pose and latest keyframe pose never jump. After an update, keyframes which have shifted must have their point clouds transformed in the odometry module; this is executed in a background thread, with submap keyframes being prioritized.

\section{Results}
\label{sec:results}

In this section, we first provide an analysis of each proposed contribution to convince the reader that our core innovations are reasonable for improving the accuracy and resiliency of localization and mapping. Then, to validate our methods and system as a whole, DLIOM's accuracy and efficiency was compared against several current state-of-the-art and open-source systems. These include three LO algorithms, namely DLO~\cite{chen2022direct}, CT-ICP~\cite{dellenbach2021cticp}, and KISS-ICP~\cite{vizzo2023kiss}, and three LIO algorithms, namely LIO-SAM~\cite{shan2020lio}, FAST-LIO2~\cite{xu2022fast}, and DLIO~\cite{chen2022dlio}.
We use the entirety of two public benchmark datasets, in addition to a self-collected dataset around a university campus, to compare the algorithms. These benchmark datasets include the Newer College dataset \cite{ramezani2020newer} and the extension to Newer College Extension dataset \cite{zhang2021multicamera}.
Note that some well-known algorithms (e.g., Wildcat~\cite{ramezani2022wildcat} and X-ICP~\cite{tuna2022x}) could not be thoroughly compared against due to closed-source implementation and/or custom unreleased datasets; however, Wildcat~\cite{ramezani2022wildcat} was briefly evaluated against using the MulRan DCC03 dataset \cite{gskim-2020-mulran} as they provide numerical results of this dataset in their manuscript. 
Finally, we  demonstrate the usage of DLIOM in a fully closed-loop flight through several aggressive autonomous maneuvers in a lab setting using our custom aerial platform.

\subsection{Analysis of Components}

\subsubsection{Slip-Resistant Keyframing}
\begin{figure}[!t]
    \centering
    \includegraphics[width=0.95\columnwidth]{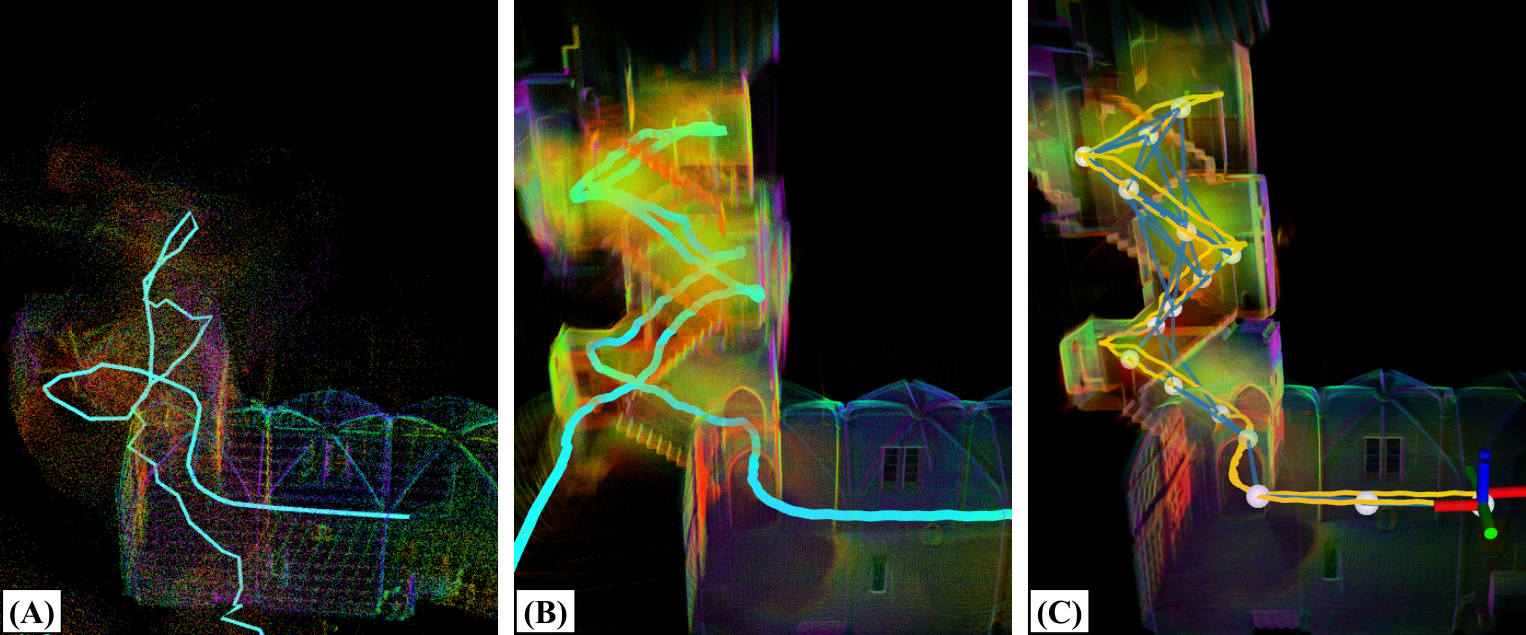}
    \caption{\textbf{Slip-Resistant Localization.} Comparison of maps and trajectories generated by (A) LIO-SAM \cite{shan2020lio}, (B) FAST-LIO2~\cite{xu2022fast}, and (C) our method, using the Newer College Extension - Stairs dataset \cite{zhang2021multicamera}. For (A), we observed slippage right after entering the stairwell, while for (B), tracking was shaking during ascension (e.g., blurry map), with it slipping after descension at the bottom. For (C), our keyframe placement (white nodes) allowed our algorithm to track sufficiently both during ascension and descension, constructing a clear map and accurate trajectory.}
    \label{fig:slip_resistant}
    \vskip -0.2in
\end{figure}

To showcase the resiliency of our slip-resistant keyframing strategy, which continually monitors scan-matching optimality and places an environmental keyframe during the onset of slippage, we use the Newer College Extension - Stairs dataset \cite{zhang2021multicamera} and compare against LIO-SAM \cite{shan2020lio} and FAST-LIO2~\cite{xu2022fast}. Staircases are notoriously difficult for SLAM algorithms---especially those which are LiDAR-centric---due to the sensors' limited field-of-view in the Z direction. Because of this, tracking can be challenging as there are less data points to associate with during ascension. This can be observed in Fig.~\ref{fig:slip_resistant}. For LIO-SAM, the algorithm slipped right at the entrance of the stairwell, most likely due to a lack in sufficient features for feature extraction that the algorithm relies on. For FAST-LIO2, tracking was much better (as the algorithm also performs no feature extraction), but localization was jittery near the apex (e.g., blurry map at the top), and the algorithm completely slipped during descension. However, by detecting the onset of slippage, DLIOM can actively place new keyframes to continually anchor itself through space and therefore allow for tracking in challenging scenarios. This can be further seen in Fig.~\ref{fig:main}(D), in which our method was able to track through eight flights of stairs, while all other algorithms failed.

\subsubsection{Jaccard Submapping}

The efficacy of our submapping strategy, which directly computes each environmental keyframe's relevancy (i.e., overlap) through a computed 3D Jaccard index and subsequently extracts those which are most useful for scan-matching, is compared against a naive, implicit method. More specifically, the naive approach extracts keyframes which are spatially nearby and those which construct the convex hull of keyframes. First proposed in \cite{chen2022direct}, this strategy implicitly assumes that these keyframes are the best for globally aligning the current scan. However, as seen in Table~\ref{results:table:jaccard}, which compares trajectory error between the naive method (``NN + Convex") and our Jaccard method (``Jaccard Index") using the Newer College Extension - Cloister dataset (Fig.~\ref{fig:submap_jaccard}), this may not extract the most relevant submap for scan-to-map alignment and can be detrimental to accuracy and computational complexity. Heuristically extracting such keyframes risks using point clouds which are not used for scan-to-map, and therefore adds unnecessary operations during kdtree or covariance structure building. In contrast, an explicit extraction of the most useful keyframes provides scan-to-map a more practical set of keyframes to align with, ultimately helping with data association for GICP and reducing computational waste on keyframes which may not be used at all for registration. This applies for all scenarios, such as ascension of a staircase, where nearby keyframes may have zero overlap but would be extracted using the nearest-neighbor method.

\begin{table}[!t]
    \centering
    \footnotesize
    \setlength{\tabcolsep}{8 pt}
    \renewcommand{\arraystretch}{1.75}
    \caption{Comparison of Submapping Strategies\cite{gskim-2020-mulran}}
    \begin{tabular}{|l||c|c|c|c|}
    \hline
   \multicolumn{1}{|c||}{\multirow{2}{*}{Submapping Strategy}} 
    & \multicolumn{3}{c|}{\multirow{1}{*}{Absolute Trajectory Error [m]}} \\ \cline{2-4}
    & Max & Mean & RMSE \\ \hline
NN + Convex \cite{chen2022direct} & 0.5898 & 0.0923 $\pm$ 0.444 & 0.1006 \\ \hdashline
Jaccard Index & \textbf{0.2613} & \textbf{0.0604} $\pm$ \textbf{0.0361} & \textbf{0.0546} \\ \hline
    \end{tabular}
    \label{results:table:jaccard}
\end{table}

\subsubsection{Connective Mapping}

Finally, we verify the effectiveness of adding connective factors between keyframes in our factor graph mapper. These factors provide additional constraints between overlapping nodes to better locally constrain each keyframe relative to one another, which can help with mapping accuracy after loop closure. We compared overall cloud-to-cloud distance to ground truth between a map generated with connectivity factors between keyframes, and a map generated only with sequential (``odometry"-like) factors. The Newer College Extension - Maths (H) dataset was used in this experiment, and ground truth was provided by a high-grade Leica BLK360 laser scanner. Cloud-to-cloud error was computed using the CloudCompare application \cite{cloudcompare} after manual registration. All clouds were voxelized with a leaf size of 0.1m to provide a fair comparison, and a maximum threshold of 1m was set for computing the average error to filter non-overlaping regions. Without connective factors, DLIOM's output map had a mean cloud-to-cloud distance to ground truth of 0.3285 $\pm$ 0.2289; however, with connective factors, this cloud-to-cloud distance reduced down to 0.2982 $\pm$ 0.2214. These connective factors between overlapping nodes can create a more accurate map after graph optimization by providing additional constraints for loop closures.

\subsubsection{Adaptive Scan-Matching}

\begin{figure}[!t]
    \centering
    \includegraphics[width=0.95\columnwidth]{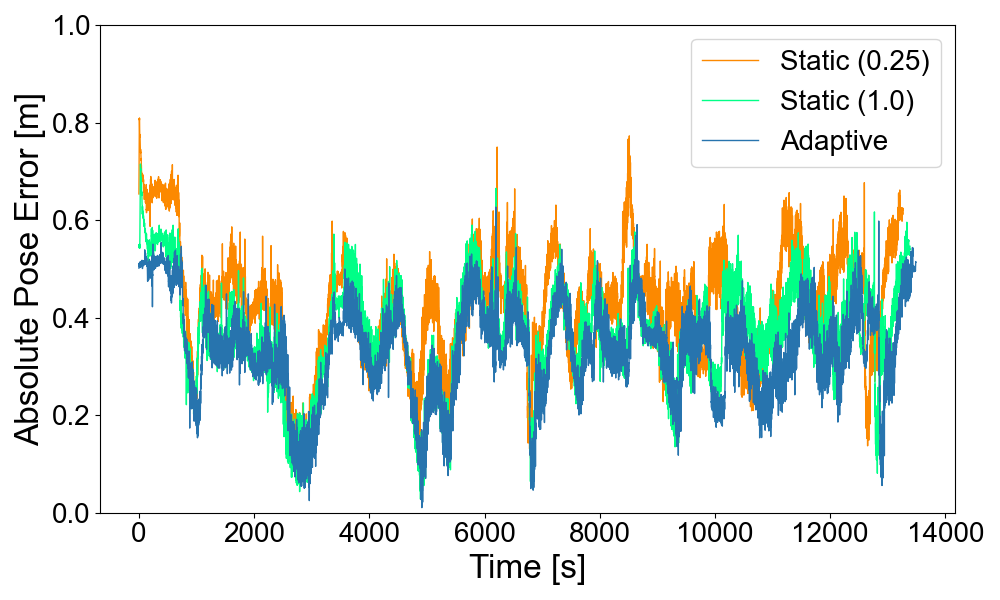}
    \caption{\textbf{Adaptive Scan-Matching.} A comparison of absolute pose error on the Newer College - Short Experiment dataset using adaptive and static scan-matching correspondence thresholds. We observed, on average, a lower trajectory error using our adaptive scaling technique as compared to static search thresholds which other methods typically use; this allows for more consistent localization in both small and large environments.}
    \label{fig:adaptive_scanmatching}
    \vskip -0.1in
\end{figure}

Next, we compared our adaptive scan-matching technique, which scales the GICP maximum correspondence distance according to scan sparsity, against two statically-set thresholds. Specifically, we compared against a static correspondence distance of $0.25$, which is typically optimal for smaller environments (since points are closer together), and to the trajectory from a static correspondence distance of $1.0$, which is more reasonable for larger, outdoor environments. We used the Newer College - Short Experiment dataset for this comparison, as it features three different sections of varying sizes (``Quad", ``Mid-Section", and ``Parkland"). The results are shown in Fig.~\ref{fig:adaptive_scanmatching}, in which we observed our adaptive thresholding scheme to perform the best, followed by a static threshold of $1.0$, and finally a threshold of $0.25$ which performed the worst amongst the three. This is reasonable because the majority of the Short Experiment dataset is in medium and large (``Quad" and ``Parkland") scenes, with about 20\% of the trajectory in the smaller ``Mid-Section." Because of this, a larger threshold would, on average, perform better than a smaller threshold (RMSE of 0.3810 $\pm$ 0.1063 versus 0.4140 $\pm$ 0.1167), but not as well as one that adapts to provide the best of both worlds (0.3571 $\pm$ 0.0971).

\subsubsection{Comparison of Motion Correction}

\begin{figure}[!t]
    \centering
    \includegraphics[width=0.95\columnwidth]{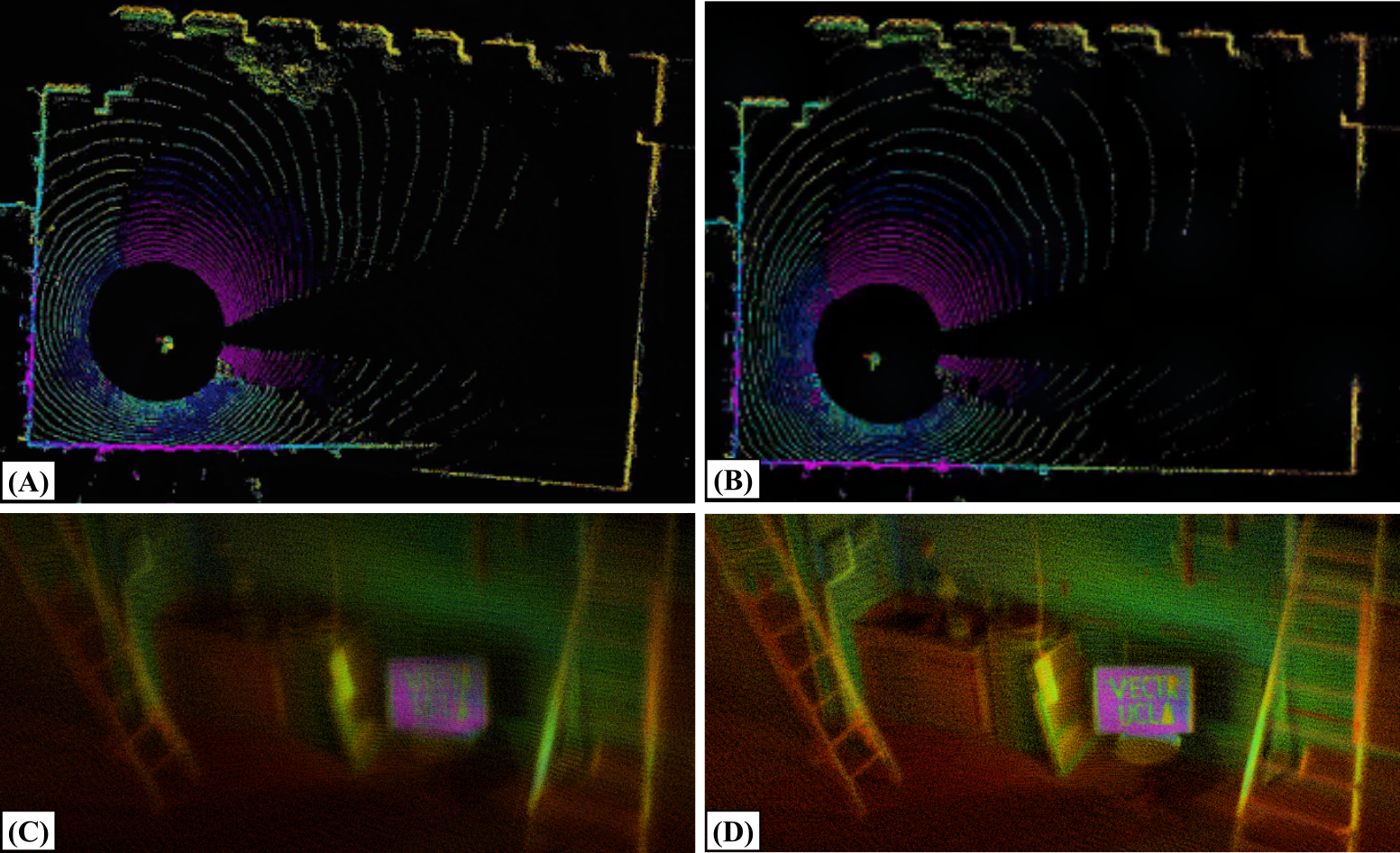}
    \caption{\textbf{Motion Correction.} Distortion caused by rapid movement can severely skew LiDAR scans which affects scan-matching (and therefore localization) accuracy. (A) A point cloud from the Newer College - Dynamic dataset with severe distortion, which causes the bottom walls to misalign. (B) A motion-corrected cloud using our coarse-to-fine scheme which accurately reconstructs the environment. This phenomenon is also observed in our method's output map, in which maps generated from aggressive maneuvers are blurrier without motion correction (C) than those generated with (D).}
    \label{fig:motioncorrection}
    \vskip -0.2in
\end{figure}

To investigate the impact of our proposed motion correction scheme, we first conducted an ablation study with varying degrees of deskewing using the Newer College dataset~\cite{ramezani2020newer}. Each of the tested algorithms employ a different degree and method of motion compensation, therefore creating an exhaustive comparison to the current state-of-the-art. To isolate our new additions since our previous work, we used DLIO~\cite{chen2022dlio} in this study, which ranged from no motion correction (None), to correction using only nearest IMU integration via (\ref {eq:deskew}) (Discrete), and finally to full continuous-time motion correction via both (\ref{eq:deskew}) and (\ref{eq:deskew_timestamp}) (Continuous) (Table~\ref{results:table:newer2020}). Particularly of note is the Dynamic dataset, which contained highly aggressive motions with rotational speeds up to 3.5 rad/s. With no correction, error was the highest among all algorithms at $0.1959$ RMSE. With partial correction, error significantly reduced due to scan-matching with more accurate and representative point clouds; however, using the full proposed scheme, we observed an error of only $0.0612$ RMSE---the lowest among all tested algorithms. With similar trends for all other datasets, the superior tracking accuracy granted by better motion correction is clear: constructing a unique transform in continuous-time creates a more authentic point cloud than previous methods, which ultimately affects scan-matching and therefore trajectory accuracy. Fig.~\ref{fig:motioncorrection} showcases this empirically: our coarse-to-fine technique constructs point clouds that more accurately represents the environment compared to methods with simple or no motion correction.

\begin{table*}[!t]
    \centering
    \footnotesize
    \setlength{\tabcolsep}{8 pt}
    \renewcommand{\arraystretch}{1.75}
    \caption{Comparison with Newer College Dataset \cite{ramezani2020newer}}
    \begin{tabular}{|l|c||c|c|c|c|c|c|}
    \hline
   \multicolumn{1}{|c|}{\multirow{3}{*}{Algorithm}} 
    & \multicolumn{1}{c||}{\multirow{3}{*}{Type}} 
    & \multicolumn{5}{c|}{\multirow{1}{*}{Absolute Trajectory Error (RMSE) [m]}} 
    & \multicolumn{1}{c|}{\multirow{3}{10mm}{\centering Average Comp. [ms]}} \\ \cline{3-7}
    & & Short Exp. & Long Exp. & Quad w/ Dynamics & Dynamic Spinning & Parkland Mount & \\ \cdashline{3-7}
    & & \textit{1609.40m} & \textit{3063.42m} & \textit{479.04m} & \textit{97.20m} & \textit{695.68m} & \\ \hline
DLO \cite{chen2022direct} & LO & 0.4633 & 0.4125 & 0.1059 & 0.1954 & 0.1846 & 48.10\\ \hdashline
CT-ICP~\cite{dellenbach2021cticp} & LO & 0.5552 & 0.5761 & 0.0981 & 0.1426 & 0.1802 & 412.27 \\ \hdashline
KISS-ICP~\cite{vizzo2023ral} & LO & 0.6675 & 1.5311 & 0.1040 & Failed & 0.2027 & 167.38 \\ \hdashline
LIO-SAM~\cite{shan2020lio} & LIO & 0.3957 & 0.4092 & 0.0950 & 0.0973 & 0.1761 & 179.33 \\ \hdashline
FAST-LIO2~\cite{xu2022fast} & LIO & 0.3775 & 0.3324 & 0.0879 & 0.0771 & 0.1483 & 42.86 \\ \hdashline
DLIO (None) & LIO & 0.4299 & 0.3988 & 0.1117 & 0.1959 & 0.1821 & 34.88 \\ \hdashline
DLIO (Discrete) & LIO & 0.3803 & 0.3629 & 0.0943 & 0.0798 & 0.1537 & \textbf{34.61} \\ \hdashline
DLIO (Continuous) \cite{chen2022dlio} & LIO & 0.3606 & 0.3268 & 0.0837 & 0.0612 & 0.1196 & 35.74 \\ \hdashline
DLIOM & LIO & \textbf{0.3571} & \textbf{0.3252} & \textbf{0.0821} & \textbf{0.0609} & \textbf{0.1181} & 36.21 \\ \hline
    \end{tabular}
    \label{results:table:newer2020}
    \vskip -0.1in
\end{table*}

\subsection{Benchmark Results}
We compare the accuracy and efficiency of DLIOM against six state-of-the-art algorithms using public and self-collected datasets. Aside from extrinsics, default parameters at the time of writing for each algorithm were used in all experiments unless otherwise noted. Specifically, loop-closures were kept enabled for LIO-SAM and online extrinsics estimation disabled for FAST-LIO2 to provide the best results of each algorithm. For FAST-LIO2, we reduced the default crop otherwise it would fail in smaller environments. Deskewing was enabled for KISS-ICP, and for CT-ICP, voxelization was increased and data playback speed was slowed down to 25\% otherwise the algorithm would fail due to significant frame drops. Loop closures were disabled in DLIOM to provide a more fair assessment. Trajectories were compared against the ground truth using evo~\cite{grupp2017evo} in TUM~\cite{sturm12iros} format and aligned with the Umeyama algorithm~\cite{umeyama} for all public benchmark datasets. Algorithms which did not produce meaningful results are indicated accordingly in the tables, and trajectory lengths for each dataset are indicated in italics to give a reader a sense of duration. All tests were conducted on a 16-core Intel i7-11800H CPU.

\subsubsection{Newer College Dataset}

\begin{figure}[!t]
    \centering
    \includegraphics[width=0.95\columnwidth]{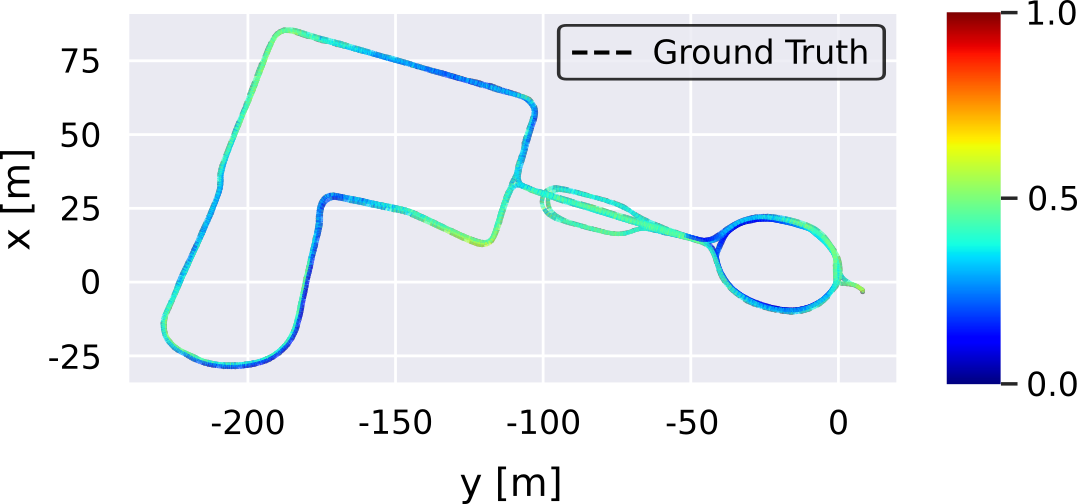}
    \caption{\textbf{Trajectory of Long Experiment.} The generated trajectory for the Newer College - Long Experiment. Color indicates absolute pose error.}
    \label{results:fig:newer2020traj}
    \vskip -0.2in
\end{figure}

Trajectory accuracy and average per-scan time of all algorithms were also compared using the original Newer College benchmark dataset~\cite{ramezani2020newer}. For these tests, we used data from the Ouster OS1-64 (10Hz) in addition to its internal IMU (100Hz) to ensure accurate time synchronization between sensors. For certain Newer College datasets, the first 100 poses were excluded from computing FAST-LIO2's RMSE due to slippage at the start in order to provide a fair comparison. 
Short, Long, and Parkland experiments were routes recorded at a standard walking pace around several different sections, while Quad and Dynamic featured rapid linear and angular movements.
The results are shown in Table~\ref{results:table:newer2020}, in which we observed our method to produce the lowest trajectory RMSE as compared to all other algorithms.
Fig.~\ref{results:fig:newer2020traj} illustrates DLIOM's low trajectory error compared to ground truth for the Newer College - Long Experiment dataset even after over three kilometers of travel.

\subsubsection{Newer College Extension Dataset}

\begin{table*}[!t]
    \centering
    \footnotesize
    \setlength{\tabcolsep}{5 pt}
    \renewcommand{\arraystretch}{1.75}
    \caption{Comparison with Newer College Extension Dataset \cite{zhang2021multicamera}}
    \begin{tabular}{|l|c||c|c|c|c|c|c|c|c|c|c|}
    \hline
   \multicolumn{1}{|c|}{\multirow{3}{*}{Algorithm}} 
    & \multicolumn{1}{c||}{\multirow{3}{*}{Type}} 
    & \multicolumn{9}{c|}{\multirow{1}{*}{Absolute Trajectory Error (RMSE) [m]}} 
    & \multicolumn{1}{c|}{\multirow{3}{10mm}{\centering Average Comp.  [ms]}} \\ \cline{3-11}
    & & Quad (E) & Quad (M) & Quad (H) & Stairs & Cloister & Park & Maths (E) & Maths (M) & Maths (H) & \\ \cdashline{3-11}
    & & \textit{246.67m} & \textit{260.36m} & \textit{234.81m} & \textit{57.04m} & \textit{428.79m} & \textit{2396.20m} & \textit{263.62m} & \textit{304.28m} & \textit{320.56m} & \\ \hline
DLO \cite{chen2022direct} & LO & 0.0866 & 0.1141 & 0.1490 & 0.1605 & 0.1608 & 0.8108 & 0.1658 & 0.7730 & 1.0864 & 31.34 \\ \hdashline
CT-ICP~\cite{dellenbach2021cticp} & LO & 0.0974 & 0.1820 & Failed & Failed & 0.3558 & 0.7935 & 0.0970 & 0.1362 & Failed & 213.41 \\ \hdashline
KISS-ICP~\cite{vizzo2023ral} & LO & 0.0960 & 0.2016 & 0.3663 & Failed & 0.7398 & 0.9631 & 0.0718 & 0.1203 & 0.2789 & 62.38 \\ \hdashline
LIO-SAM~\cite{shan2020lio} & LIO & 0.0714 & 0.0718 & 0.0909 & Failed & 0.0811 & 0.8371 & 0.0784 & 0.1168 & 0.0932 & 51.57 \\ \hdashline
FAST-LIO2~\cite{xu2022fast} & LIO & 0.0491 & 0.0608 & 0.0670 & Failed & 0.0594 & 0.2678 & 0.0872 & 0.1024 & 0.0646 & 28.04 \\ \hdashline
DLIO~\cite{chen2022dlio} & LIO & 0.0388 & 0.0610 & 0.0631 & 0.1559 & 0.0855 & 0.2866 & 0.0786 & 0.0983 & 0.0863 & \textbf{25.99} \\ \hdashline
DLIOM & LIO & \textbf{0.0350} & \textbf{0.0571} & \textbf{0.0615} & \textbf{0.0686} & \textbf{0.0546} & \textbf{0.2608} & \textbf{0.0609} & \textbf{0.0904} & \textbf{0.0609} & 27.89 \\ \hline
    \end{tabular}
    \label{results:table:newer2021}
    \vskip -0.1in
\end{table*}

Additionally, we compared all algorithms using the latest extension to the Newer College benchmark dataset~\cite{zhang2021multicamera}, which features three collections of data. Collections 1 and 3 contains three datasets each with progressively increasing difficulty, from ``Easy" (E), which had slow paced movement, to ``Medium" (M), which featured slightly more aggressive turn-rates and motions, and finally to ``Hard" (H), which contained highly aggressive motions, rotations, and locations in both small and large environments. Collection 2 contains three datasets, each of which are highly different than the other to create a diverse set of environments. This includes traversing up and down a staircase, walking around a cloister with limited visibility, and a large-scale park with multiple loops. In this benchmark dataset, we used data from the Ouster OS0-128 (10Hz) in addition to the Alphasense Core IMU (200Hz), since this particular extension had well-synchronized data.

The results are shown in Table~\ref{results:table:newer2021} for all tested algorithms. Particularly of note are the two Hard (H) datasets, in addition to the Stairs dataset. For Quad (H) and Maths (H), both of which had highly aggressive and unpredictable movements, CT-ICP failed (even when reducing playback speed down to 10\%), while both DLO and KISS-ICP had significantly higher trajectory errors. This demonstrates the strength and need for fusing inertial measurement units for point cloud motion correction. For Stairs, most algorithms failed to produce meaningful results due to the difficult ascension and the limited vertical field-of-view from the LiDAR sensor. Of those which could track sufficiently, both DLO and DLIO had significantly high errors; however, by detecting and placing a new keyframe right at the onset of slippage, DLIOM is able to achieve a low RMSE of just $0.0686$m. This is further illustrated in Fig.~\ref{fig:slip_resistant}, in which keyframes (white nodes) are placed at locations with high scene change (e.g., through the door, in-between stairs).

\subsubsection{MulRan Dataset}

\begin{table}[!t]
    \centering
    \footnotesize
    \setlength{\tabcolsep}{8 pt}
    \renewcommand{\arraystretch}{1.75}
    \caption{Comparison with MulRan DCC03\cite{gskim-2020-mulran}}
    \begin{tabular}{|l||c|c|c|}
    \hline
   \multicolumn{1}{|c||}{\multirow{2}{*}{Algorithm}} 
    & \multicolumn{2}{c|}{\multirow{1}{*}{Relative Trajectory Error (RPE)}} \\ \cline{2-4}
    & Translation [\%] & Rotation [\textdegree/m] \\ \hline
LIO-SAM~\cite{shan2020lio} & \textbf{2.4} & 0.009 \\ \hdashline
FAST-LIO2~\cite{xu2022fast} & 6.8 & 0.030 \\ \hdashline
Wildcat \cite{ramezani2022wildcat} & 2.9 & 0.010 \\ \hdashline
DLIOM & \textbf{2.4} & \textbf{0.007} \\ \hline
    \end{tabular}
    \label{results:table:mulran}
\end{table}

\begin{figure}[!t]
    \centering
    \includegraphics[width=0.95\columnwidth]{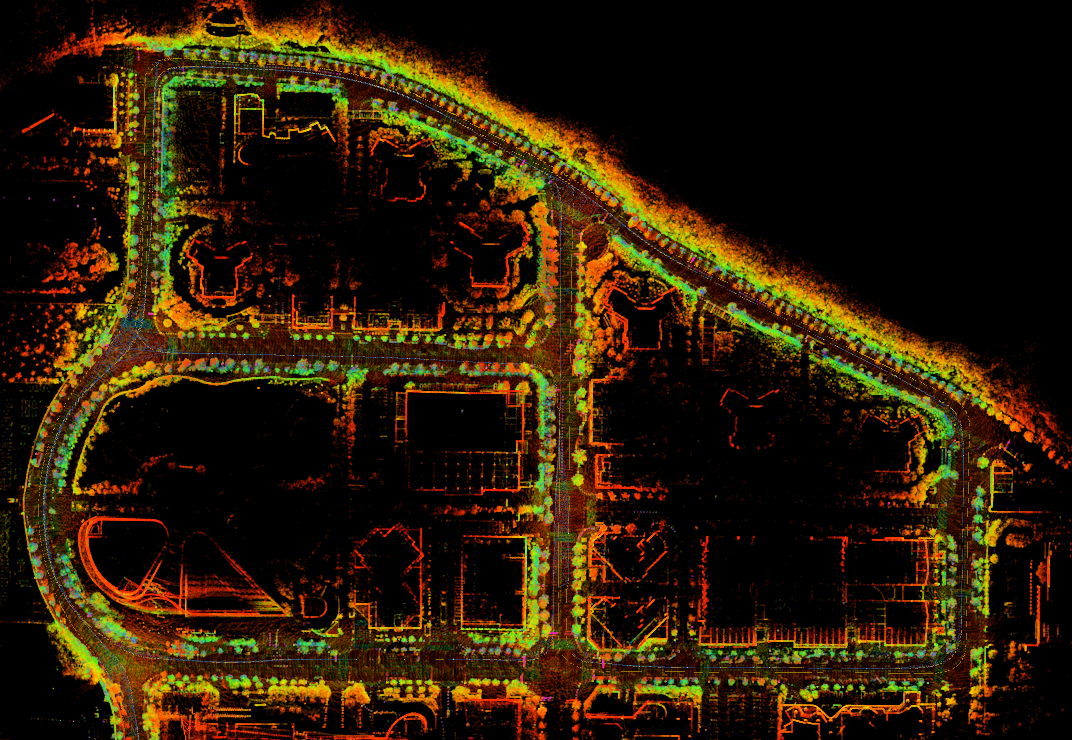}
    \caption{\textbf{MulRan DCC03.} Top-down view of the map of the MulRan DCC03 dataset, generated by DLIOM. This specific dataset featured approximately 5421.82 meters of travel from driving around three different loops in Korea.}
    \label{fig:mulran}
    \vskip -0.2in
\end{figure}

To compare against Wildcat~\cite{ramezani2022wildcat}, we use the MulRan DCC03~\cite{gskim-2020-mulran} dataset. This is shown in Table~\ref{results:table:mulran} (results for LIO-SAM, FAST-LIO2, and Wildcat retrieved from \cite{ramezani2022wildcat}). While the details of specifically how the relative trajectory error (RPE) was computed are unclear, we assume that the translational metric was the average RPE with respect to the point distance error ratio using evo\cite{grupp2017evo}, and the rotational metric was using evo's ``rot\_part" option. A fair comparison of absolute trajectory error could not be conducted, as numerical values were not provided by the authors, but DLIOM's ATE was on average 2.36m and 2.4\textdegree. Top-down map is shown in Fig~\ref{fig:mulran}.

\subsubsection{UCLA Campus Dataset}

\begin{table*}[!t]
    \centering
    \footnotesize
    \setlength{\tabcolsep}{8 pt}
    \renewcommand{\arraystretch}{1.75}
    \caption{Comparison with UCLA Campus Dataset}
    \begin{tabular}{|l|c||c|c|c|c|c|c|c|c|}
        \hline
        \multicolumn{1}{|c|}{\multirow{3}{*}{Algorithm}} &
        \multicolumn{1}{c||}{\multirow{3}{*}{Type}} &
          \multicolumn{2}{c|}{Royce Hall (A)} & 
          \multicolumn{2}{c|}{Court of Sciences (B)} &
          \multicolumn{2}{c|}{Bruin Plaza (C)} &
          \multicolumn{2}{c|}{Sculpture Garden (D)} \\ \cdashline{3-10}
          & & 
          \multicolumn{2}{c|}{\textit{652.66m}} & 
          \multicolumn{2}{c|}{\textit{526.58m}} &
          \multicolumn{2}{c|}{\textit{551.38m}} &
          \multicolumn{2}{c|}{\textit{530.75m}} \\ \cline{3-10}
          & & Error [m] & Comp [ms] & Error [m] & Comp [ms] & Error [m] & Comp [ms] & Error [m] & Comp [ms] \\ \hline
          
        DLO~\cite{chen2022direct} & LO & 0.0216 & 20.40 & 1.2932 & 20.77 & 0.0375 & 21.18 & 0.0178 & 21.62 \\ \hdashline
        CT-ICP~\cite{dellenbach2021cticp} & LO & 0.0387 & 351.85 & 0.0699 & 342.76 & 0.0966 & 334.15 & 0.0253 & 370.19 \\ \hdashline
        KISS-ICP~\cite{vizzo2023ral} & LO & 0.0689 & 50.28 & 0.4007 & 31.84 & 0.2412 & 35.31 & 0.0987 & 62.96 \\ \hdashline
        LIO-SAM~\cite{shan2020lio} & LIO & 0.0216 & 33.21 & 0.0692 & 29.14 & 0.0936 & 39.04 & 0.0249 & 48.94 \\ \hdashline
        FAST-LIO2~\cite{xu2022fast} & LIO & 0.0454 & 15.39 & 0.0353 & 12.25 & 0.0363 & 14.84 & 0.0229 & 15.01 \\ \hdashline
        DLIO~\cite{chen2022dlio} & LIO & 0.0105 & \textbf{10.45} & 0.0233 & \textbf{8.37} & 0.0301 & \textbf{8.66} & 0.0082 & \textbf{10.96} \\ \hdashline
        DLIOM & LIO & \textbf{0.0097} & 12.91 & \textbf{0.0228} & 14.18 & \textbf{0.0235} & 15.37 & \textbf{0.0078} & 13.78 \\ \hline
    \end{tabular}
    \label{results:table:ucla}
\end{table*}

\begin{figure*}[!t]
    \centering
    \includegraphics[width=0.99\textwidth]{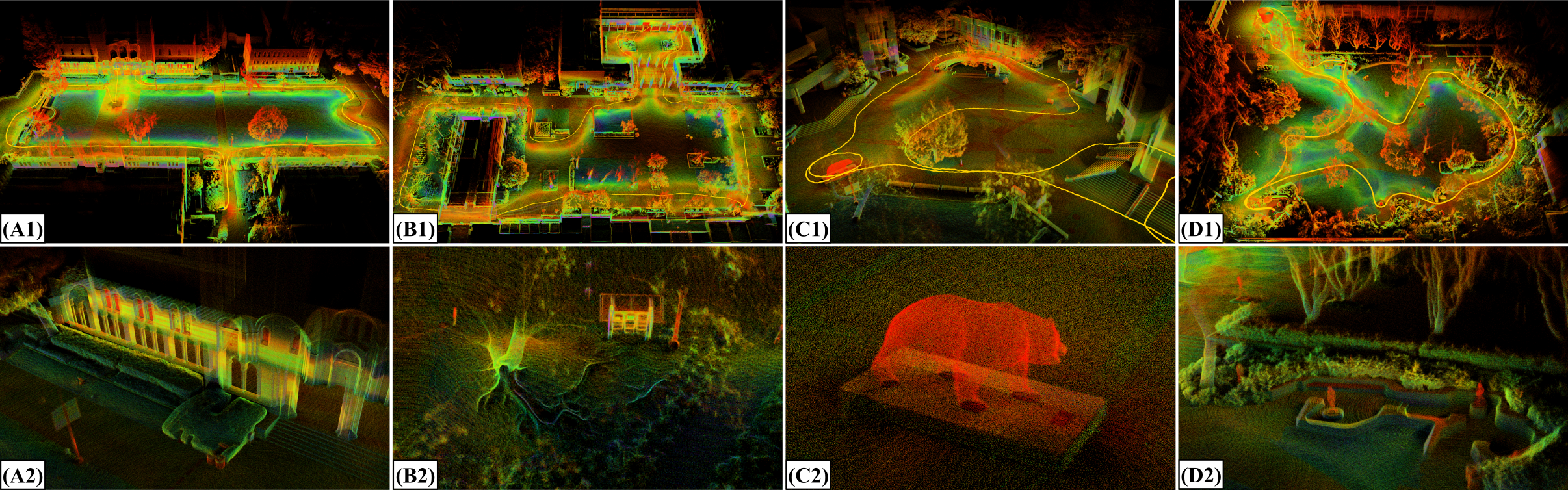}
    \caption{\textbf{UCLA Campus.} Detailed maps of locations around UCLA in Los Angeles, CA generated by our algorithm, including (A) Royce Hall in Dickson Court, (B) Court of Sciences, (C) Bruin Plaza, and (D) the Franklin D. Murphy Sculpture Garden, with both (1) a bird's eye view and (2) a close-up to demonstrate the level of fine detail DLIO can generate. The trajectory taken to generate these maps is shown in yellow in the first row.}
    \label{fig:ucla_maps}
    \vskip -0.2in
\end{figure*}

We additionally showcase our method's accuracy using four large-scale datasets at UCLA for additional comparison (Fig.~\ref{fig:ucla_maps}). These datasets were gathered by hand-carrying our aerial platform (Fig.~\ref{fig:main}) over 2261.37m of total trajectory. Our sensor suite included an Ouster OS1 (10Hz, 32 channels recorded with a 512 horizontal resolution) and a 6-axis InvenSense MPU-6050 IMU located approximately 0.1m below it. We note here that this IMU can be purchased for approximately \$10, demonstrating that LIO algorithms need not require high-grade IMU sensors that previous works have used. Note that a comparison of absolute trajectory error was not possible due to the absence of ground truth, so as is common practice, we compute end-to-end translational error as a proxy metric (Table~\ref{results:table:ucla}). In these experiments, our method outperformed all others across the board in end-to-end translational error. However, similar to the trends found in the Newer College datasets, our average per-scan computational time has slightly increased due to the new algorithmic additions since DLIO. Regardless however, our resulting maps can capture fine detail in the environment which ultimately provides more intricate information cues for autonomous mobile robots such as terrain traversability.


\section{Discussion}
\label{sec:discussion}

This letter presents Direct LiDAR-Inertial Odometry and Mapping (DLIOM), a robust SLAM algorithm with an extreme focus on operational reliability and accuracy to yield real-time state estimates and environmental maps across a diverse set of domains. DLIOM mitigates several common failure points in typical LiDAR-based SLAM solutions through an architectural restructuring and several algorithmic innovations. Rather than using a single sensor fusion framework (e.g., probabilistic filter or graph optimization) to produce both localization and map as is typical in other algorithms, we separate these two processes into separate threads and tackle them independently. Leveraging a nonlinear geometric observer guarantees the convergence of IMU propagation towards LiDAR scan-matching and reliably initializes velocity and sensor biases, which is required for our fast coarse-to-fine motion correction technique. On the other hand, a factor graph, with nodes at keyframe locations determined by our odometry thread, continually optimizes for a best-fit map using connective factors between overlapping keyframes, which provide extra relative constraints to the optimization problem.

Fast and robust localization is achieved hierarchically in the front-end's scan-matching, keyframing and submapping processes. An adaptive scan-matching method automatically tunes the maximum distance between corresponding planes for GICP by computing a novel point cloud sparsity metric, resulting in more consistent registration in differently sized environments. Slip-resistant keyframing ensures a sufficient number of data correspondences between the scan and the submap by detecting abrupt scene changes using a new sensor-agnostic degeneracy metric. Finally, our submap is explicitly generated by computing the 3D Jaccard index between the current scan and each environmental keyframe to ensure maximal overlap in the submap for data correspondence searching. These ideas collectively enable a highly reliable LiDAR SLAM system that is not only agnostic to the operating environment, but is also fast and online for real-time usage on computationally-constrained platforms.

\bibliographystyle{IEEEtran}
\bibliography{references}

\end{document}